\documentclass[5p,final,authoryear,twocolumn]{elsarticle}
\usepackage[utf8]{inputenc}
\usepackage{graphicx}
\usepackage{amssymb}

\usepackage{array}
\usepackage{amsmath}
\usepackage{hyperref}
\usepackage{url}
\usepackage{multirow}
\usepackage{booktabs}
\usepackage{caption}
\usepackage{subcaption}

\newcommand{\otoprule}{\midrule[\heavyrulewidth]}

\def\eg{{\em e.g.}}
\def\ie{{\em i.e.}}
\def\vs{{\em vs.}}

\newcommand{\cls}{y}

\newcommand{\obs}{\mathbf{x}}

\newcommand{\SetOfClasses}{\mathcal{Y}}
\newcommand{\SetOfInlierClasses}{\mathcal{Y}_{\text{in}}}
\newcommand{\SetOfOutlierClasses}{\mathcal{Y}_{\text{out}}}
\newcommand{\SetOfTrainOutlier}{\mathcal{Y}_{\text{out}}^{\text{train}}}
\newcommand{\SetOfValOutlier}{\mathcal{Y}_{\text{out}}^{\text{val}}}
\newcommand{\SetOfTestOutlier}{\mathcal{Y}_{\text{out}}^{\text{test}}}

\newcommand{\dataset}{\mathcal{D}}
\newcommand{\datasetTrain}{\mathcal{D}^{\text{train}}}
\newcommand{\datasetVal}{\mathcal{D}^{\text{val}}}
\newcommand{\datasetTest}{\mathcal{D}^{\text{test}}}
\newcommand{\datasetinlier}{\mathcal{D}_{\text{in}}}
\newcommand{\datasettraininlier}{\mathcal{D}_{\text{in}}^{\text{train}}}
\newcommand{\datasetvalinlier}{\mathcal{D}_{\text{in}}^{\text{val}}}
\newcommand{\datasettestinlier}{\mathcal{D}_{\text{in}}^{\text{test}}}
\newcommand{\datasetoutlier}{\mathcal{D}_{\text{out}}}
\newcommand{\datasettrainoutlier}{\mathcal{D}_{\text{out}}^{\text{train}}}
\newcommand{\datasetvaloutlier}{\mathcal{D}_{\text{out}}^{\text{val}}}
\newcommand{\datasettestoutlier}{\mathcal{D}_{\text{out}}^{\text{test}}}

\newcommand{\argmax}{\operatornamewithlimits{argmax}}

\begin{document}

\begin{frontmatter}


\title{Does Your Dermatology Classifier Know What It Doesn't Know? \\ Detecting the Long-Tail of Unseen Conditions}


\makeatletter
\def\@author#1{\g@addto@macro\elsauthors{\normalsize%
    \def\baselinestretch{1}%
    \upshape\authorsep#1\unskip\textsuperscript{%
      \ifx\@fnmark\@empty\else\unskip\sep\@fnmark\let\sep=,\fi
      \ifx\@corref\@empty\else\unskip\sep\@corref\let\sep=,\fi
      }%
    \def\authorsep{\unskip,\space}%
    \global\let\@fnmark\@empty
    \global\let\@corref\@empty  
    \global\let\sep\@empty}%
    \@eadauthor={#1}
}
\makeatother

\author[]{Abhijit~Guha~Roy$^1$\corref{cor2}}
\author[]{Jie~Ren$^2$\corref{cor2}}
\author[]{Shekoofeh~Azizi$^1$}
\author[]{Aaron~Loh$^1$}
\author[]{Vivek~Natarajan$^1$}
\author[]{Basil~Mustafa$^2$}
\author[]{Nick~Pawlowski$^1$}
\author[]{Jan~Freyberg$^1$}
\author[]{Yuan~Liu$^1$}
\author[]{Zach~Beaver$^1$}
\author[]{Nam~Vo$^1$}
\author[]{Peggy~Bui$^1$}
\author[]{Samantha~Winter$^1$}
\author[]{Patricia~MacWilliams$^1$}
\author[]{Greg~S.~Corrado$^1$}
\author[]{Umesh~Telang$^1$}
\author[]{Yun~Liu$^1$}
\author[]{Taylan~Cemgil$^3$}
\author[]{Alan~Karthikesalingam$^1$}
\author[]{\\ Balaji~Lakshminarayanan$^2$\corref{cor1}}
\author[]{Jim Winkens$^1$\corref{cor1}}
\cortext[cor2]{{A. Guha Roy and J. Ren are joint first authors.}}
\cortext[cor1]{{J. Winkens and B. Lakshminarayanan are joint last authors. \\
Correspondence to: 
\{agroy, jjren, balajiln, jimwinkens\}@google.com }}

\address{$^1$Google Health, $^2$Google Research, Brain Team, $^3$DeepMind}

\begin{abstract}
Supervised deep learning models have proven to be highly effective in classification of dermatological conditions. These models rely on the availability of abundant labeled training examples. However, in the real-world, many dermatological conditions are individually too infrequent for per-condition classification with supervised learning. Although individually infrequent, these conditions may collectively be common and therefore are clinically significant in aggregate. To prevent models from generating erroneous outputs on such examples, there remains a considerable unmet need for deep learning systems that can better detect such infrequent conditions. These infrequent `outlier' conditions are seen very rarely (or not at all) during training. In this paper, we frame this task as an out-of-distribution (OOD) detection problem. We set up a benchmark ensuring that outlier conditions are disjoint between the model training, validation, and test sets.
Unlike traditional OOD detection benchmarks where the task is to detect dataset distribution shift, we aim at the more challenging task of detecting subtle semantic differences.
We propose a novel hierarchical outlier detection (HOD) loss, which assigns multiple abstention classes corresponding to each training outlier class and jointly performs a coarse classification of inliers \vs{} outliers, along with fine-grained classification of the individual classes. We demonstrate that the proposed HOD loss based approach outperforms leading methods that leverage outlier data during training. Further, performance is significantly boosted by using recent representation learning methods (BiT, SimCLR, MICLe).
Further, we explore ensembling strategies for OOD detection and propose a diverse ensemble selection process for the best result.
We also perform a subgroup analysis over conditions of varying risk levels and different skin types to investigate how OOD performance changes over each subgroup and demonstrate the gains of our framework in comparison to baseline.
Furthermore, we go beyond traditional performance metrics and introduce a cost matrix for model trust analysis to approximate downstream clinical impact. We use this cost matrix to compare the proposed method against the baseline, thereby making a stronger case for its effectiveness in real-world scenarios.


\end{abstract}

\begin{keyword}
Deep learning \sep dermatology \sep ensembles \sep long-tailed recognition \sep out-of-distribution detection \sep outlier exposure \sep representation learning.


\end{keyword}

\end{frontmatter}

\section{Introduction}
\label{sec:intro}

Deep learning has been used to approximate the performance of clinicians in a plethora of clinically-meaningful classification tasks in medical imaging~\citep{xiaoliureviewarticle}, with regulatory approval for hundreds of systems to be used in clinical care~\citep{muehlematter2021approval}.
Whereas most such systems use supervised learning to perform binary classification tasks (\eg{} the presence/absence of a pathology), real clinical settings often require recognition of rarer entities in a `long tail' distribution of many possible conditions.
While a few conditions in the distribution may be sufficiently common to enable supervised training of per-condition classification models, the long tail usually comprises a significantly greater number of `outlier' conditions: those conditions that are individually too infrequent for classification using supervised learning to be practical~\citep{zhou2020review}.

\begin{figure*}[ht]
\center{\includegraphics[trim={0 0 0 0},clip,width=\textwidth]{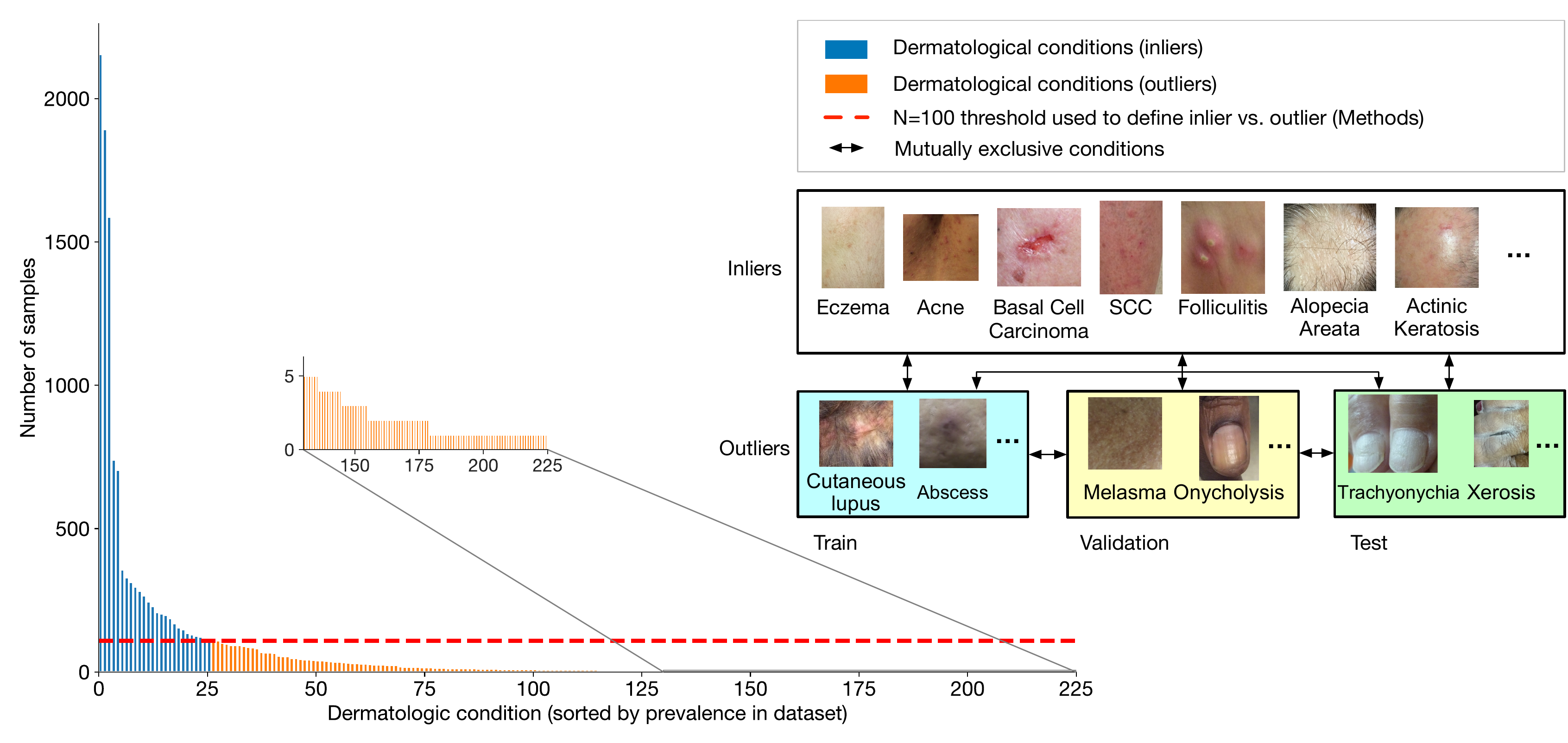}}
\caption{The figure illustrates the `long tail' distribution of different dermatological conditions in our dataset. The $26$ inlier conditions (with at least 100 samples similar to \citet{liu2020deep}) are indicated in blue. The remaining $199$ rare outlier conditions are indicated in orange. As shown in the zoomed inset of the long tail, outlier conditions can have as low as one sample per condition. We also show that the inlier and outlier conditions are mutually exclusive. The outliers are further split into train, validation and test splits with mutually exclusive conditions as indicated in the figure. The details of the splitting strategy is provided in Sec.~\ref{subsec:data_split}.}
\label{fig:graphAbs}
\end{figure*}

In other words, clinically safe performance requires that classifiers should not only achieve high accuracy on independent and identically distributed inputs (\ie{} inliers), but also reliably detect outlier inputs (which may confuse classifiers, making erroneous predictions) that do not belong to any of the classes encountered during training. 
Developing such robust and reliable classifiers could improve safety in clinical use by flagging when a condition was present that the model had not encountered during training. This information could be used to trigger a variety of practical safeguards such as having the model abstain from making a decision and instead defer to a clinician.

Classification of dermatological conditions through deep learning techniques is a typical clinical application where the long-tailed distribution of previously-unseen outliers poses a challenge~\citep{prabhu2018prototypical}.
In previous work,~\cite{liu2020deep} demonstrated that a deep learning system could effectively distinguish among $26$ of the most common skin conditions. These represent around $80\%$ of cases in the tele-dermatology dataset used.
The remaining $20\%$ of cases featured a long-tail distribution with hundreds of skin conditions that occurred considerably less frequently. 
This long-tailed distribution is illustrated in Fig.~\ref{fig:graphAbs}, with the common conditions indicated in blue and the rare ones in orange. 
Training models for classification of these rare conditions is prohibitively challenging due to the scarcity of available per-condition training examples. 

Possible solutions include using class balancing techniques or few-shot learning approaches~\citep{pmlr-v136-weng20a, prabhu2018prototypical}. However,~\cite{pmlr-v136-weng20a} showed that such approaches do not boost the performance on these rare conditions significantly, which suggests that these approaches cannot be deemed acceptable solutions for real-world deployment. 
Furthermore, the above solutions require the rare conditions to have been seen during training, whereas new unseen conditions may be encountered in real-world settings.

Another possible framing is to detect such rare conditions during test time - posing this challenge as an out-of-distribution (OOD) detection problem~\citep{bulusu2020anomalous}.
Such approaches generally leverage model confidence for this purpose. 
It has been shown that deep learning models often produce over-confident predictions for OOD inputs~\citep{goodfellow2014explaining, nguyen2015deep} suggesting that there remains room for improvement.

In this article, we aim to address the challenge of reliably detecting OOD rare dermatological conditions that were not seen during training. 
Most OOD detection works only use inlier samples during training
\citep{hendrycks2016baseline, liang2017enhancing,lakshminarayanan2016simple,lee2018simple}.
In contrast, we have access to some `known outlier' samples during training and we want to leverage them to aid detection of `unknown outlier' samples during test time.  
This is similar to setup used in~\citet{hendrycks2018deep, thulasidasan2021a}, referred to as outlier exposure, which has shown to be more effective for OOD detection.  
Furthermore, most traditional OOD benchmarks in computer vision aim at detecting distribution shifts between datasets. 
Our task aims at detecting semantic shifts (difference in semantic information for images of two different classes) between conditions, referred to as the \emph{near-OOD detection} problem~\citep{winkens2020contrastive}. 
This is more challenging as semantic distribution shifts are more subtle in comparison to dataset distribution shifts, and are thus harder to detect.

The key contributions of this article are:
\begin{itemize}
    \item We propose a novel hierarchical outlier detection (HOD) loss, and show that this outperforms existing outlier exposure based techniques for detecting OOD inputs.
    \item We introduce a near-OOD benchmarking framework and the key design choices needed for proper validation of OOD detection algorithms.
    \item We demonstrated the added utility of the novel HOD loss in the context of multiple different state-of-the-art representation learning methods (self-supervised contrastive pre-training based SimCLR and MICLe). We also show the OOD detection performance gains on large scale standard benchmarks (ImageNet and BiT model pre-trained on a large-scale JFT dataset).
    \item We propose to use a diverse ensemble with different representation learning and objectives for improved OOD detection performance. We demonstrate its superiority over vanilla ensembles and performed analysis investigating how diversity aids in better OOD detection performance.
    \item We propose a cost-weighted evaluation metric for model trust analysis that incorporates the downstream clinical implications to aid assessment of real-world impact.
\end{itemize}

We discuss related work in Sec.~\ref{sec:realted_work}, detail the benchmark setup and problem formulation in Sec.~\ref{sec:benchmark_prob}, and present our proposed method in Sec.~\ref{sec:methods}. Finally we show the experimental results in Sec.~\ref{sec:exp_res}, discuss the results in Sec.~\ref{sec:discuss} and summarize our findings in Sec.~\ref{sec:conc}.

\section{Related Work}
\label{sec:realted_work}
In this section, we present an overview of relevant recent works in OOD detection and specifically applications to medical imaging and dermatology.

\subsection{OOD Detection for Deep Learning}
In the deep learning literature, the max-of-softmax probability (MSP)~\citep{hendrycks2016baseline} is a widely-used baseline method for OOD detection due to its simplicity and good performance. As the name suggests, MSP is defined as the maximum of the predictive class probabilities from the model.
The method is based on the assumption that supervised training produces models that are less confident on OOD inputs.
To enforce low confidence predictions for OOD data, one approach is to introduce a temperature hyper-parameter to the softmax layer to decrease the confidence of classification decisions. This hyper-parameter is tuned on the validation set containing outliers~\citep{platt2000, liang2017enhancing}. 
Though in most applications, OOD data may not be available during training or validation, some medical applications are exceptions where some rare conditions may be available. 
To exploit such additional outlier data during training, methods such as Outlier Exposure (OE)~\citep{hendrycks2018deep} can be used. 
The key idea of OE is to include an extra term in the training objective for the OOD training data, additive to the regular cross entropy loss. This extra term forces a model to produce an output that is close to the uniform distribution for OOD samples such that the MSP score is lower. 
\citet{hafner2019noise} propose a similar idea for regression tasks where they force the model's predictive distribution for OOD data to be close to a prior uncertainty distribution by minimizing their KL divergence. 
Another common approach to utilize OOD training data is to add an extra, $K+1$'th OOD abstention class next to the $K$ inlier classes~\citep{thulasidasan2021a, zhang2017universum, ren2019likelihood}. 

OOD detection is conceptually related to estimation of the confidence or uncertainty in classification decisions. 
Substantial improvement in uncertainty estimation is shown to be achieved using deep ensembles~\citep{lakshminarayanan2016simple, ovadia2019can}. Deep Ensembles involves training multiple models with randomly initialized network weights and with randomly shuffled training inputs. The MSP of the average predictive class probabilities over all models is used as an confidence score. Empirically, improvements over vanilla ensembling were achieved when models trained using different hyperparameter settings (such as learning rate schedules or weight decay) or different model architectures are combined within an ensemble~\citep{wenzel2020hyperparameter, kamnitsas2017ensembles}. Though this method is known to enhance ensemble diversity and improve inlier prediction accuracy, its effect on OOD detection has not previously been explored.

More recent techniques such as pre-training, data augmentation, and self-supervised learning that enhance model robustness and generalization have also been shown to improve OOD detection performance~\citep{hendrycks2020pretrained, hendrycks2019using, hendrycks2020many,hendrycks2019augmix, venkatakrishnan2020self, winkens2020contrastive}.
In particular,~\citet{winkens2020contrastive} show that using the contrastive self-supervised training technique, SimCLR~\citep{chen2020simple}, significantly helps near-OOD detection performance. 
Using a set of class-preserving transformations, SimCLR maximizes the similarity in learned embeddings between images that are transformed from the same original image, and minimize the similarity between images that are transformed from different images. This is thought to encourage the model to learn robust and invariant features which might lead to a better OOD detection.

In addition to MSP, another test statistic that is used for OOD detection is Mahalanobis distance based OOD scoring~\citep{lee2018simple, ccalli2019frodo}. In contrast to network probability outputs, it leverages intermediate layer activations. A class conditional Gaussian distribution is fitted on the activations using training inlier data. The Mahalanobis distance between a test sample and the fitted distribution is used as an OOD score.

Another set of popular approaches is using generative models to directly fit inlier training data and use the likelihood of the test input as the OOD score. 
Several similar methods such as likelihood ratio~\citep{ren2019likelihood}, DoSE~\citep{morningstar2021density}, variational autoencoders~\citep{thiagarajan2020calibrating}, and hybrid flows~\citep{zhang2020hybrid} show promising performance on OOD detection but involve significant modifications to the training procedure and hyperparameter tuning.

There are several limitations to the existing methods discussed above. First, most of these generic OOD detection methods are evaluated using only standard benchmark datasets such as CIFAR-10, CIFAR-100, SVHN etc. While these approaches are viable for prototyping, they tend to require vast amounts of inlier training data and their OOD detection performance still needs careful evaluation in the context of each specific application, particularly for challenging settings such as medical imaging.
Second, most of the methods with the exception of~\citet{hendrycks2018deep} and~\citet{thulasidasan2021a}, do not use outlier data in the training process. In many real medical applications, some `known outlier' samples are accessible. We believe incorporating them into the training process should help to improve the performance for detecting `unknown outliers'.

\subsection{OOD Detection in Medical Imaging}
Dealing with OOD inputs is a common problem that is faced across a broad spectrum of medical imaging applications. Recent studies have investigated this challenge for chest X-ray~\citep{cao2020benchmark, ccalli2019frodo, shi2021chexseen}, brain CT scans~\citep{venkatakrishnan2020self}, fundus eye images~\citep{cao2020benchmark} and histology images~\citep{cao2020benchmark, linmans2020efficient}. 
Due to the unique challenging properties of a long-tailed distribution comprising multiple conditions, OOD detection in the dermatology setting has drawn significant attention from the research community~\citep{pacheco2020out, li2020out, combalia2020uncertainty, yasin2020open, thiagarajan2020calibrating}. 
However, current studies have several limitations. 
Firstly, most of the studies for OOD detection in dermatology tackle dermatoscopic image classification on pigmented skin lesions using standard datasets such as the ISIC challenge dataset~\citep{pacheco2020out, combalia2020uncertainty, li2020out, thiagarajan2020calibrating}. Such studies can be limited in wider clinical utility, as images need to be acquired by a special device called dermatoscope, which are typically not available outside of dermatology clinics. In addition, only a handful of pigmented skin lesion conditions are addressed in these studies, whereas in real life hundreds or thousands more skin conditions such as rashes, hair loss, and nail conditions exist and may occur more frequently. Images in those datasets are also well-lit and magnified on the pathological region without background variability, making the classification and the OOD tasks relatively easy. 
Secondly, most of the existing methods either use different post-processing methods or complex density models. \citet{pacheco2020out} and~\citet{li2020out} leverage feature maps from a pre-trained model for OOD detection. \citet{thiagarajan2020calibrating} used a variational autoencoder to learn a disentangled latent representation to improve model interpretability and designed a calibration-driven learning approach to produce a prediction interval for uncertainty quantification. However, the method requires extensive modifications to the original classification model and very careful hyperparameter tuning. 
Thirdly, most of these methods focus mainly on the comparatively easy task of far-OOD detection: non-dermatology images or poor-quality images. A more challenging task is detecting previously unseen dermatological conditions, which is a near-OOD problem and remains relatively unexplored.

\section{Benchmark Setup and Problem Formulation}
\label{sec:benchmark_prob}
\subsection{Dataset Description} 
\label{sec:dataset_description}
In this article, we use a subset of the de-identified dataset used in~\cite{liu2020deep} for model development. Cases in this dataset were collected from $17$ different sites across California and Hawaii.
Each case in the dataset consists of up to $6$ RGB images, taken by medical assistants using consumer-grade digital cameras. The images exhibit a large amount of variation in terms of affected anatomic location, background objects, resolution, perspective and lighting. We resized each image to $448\times448$ pixels for our training.
The dataset consists of $14,427$ cases, with the ground truth generated by aggregating the diagnoses from multiple US or Indian board certified dermatologists. The detailed annotation process is presented in~\cite{liu2020deep}.
Note that for simplicity, we removed the cases that had multiple conditions or that had multiple primary diagnoses in the ground truth.

We use an additional unlabeled dermatology dataset consisting of $271,433$ images from $114,849$ cases and $46,728$ patients for contrastive training based representation learning outlined in Sec.~\ref{subsec:rl} similar to~\citet{azizi2021big}. These images primarily come from skin cancer clinics in Australia and New Zealand, spread across $35$ different sites and $4$ Australian states. 
The  distribution of conditions is skewed towards cancerous conditions but no labels were available for them. 

\subsection{Data Splitting Strategy}
\label{subsec:data_split}
The long-tailed distribution of our dataset is illustrated in Fig.~\ref{fig:graphAbs}. Consistent with prior work~\citep{liu2020deep}, we select the $26$ most common conditions (based on primary diagnosis of each case, with number of cases with at least $100$ per condition) and consider them as inliers (indicated as blue). The remaining $199$ conditions (indicated in orange) are deemed as outliers. By definition the inlier and outlier conditions are mutually exclusive.

For setting up our OOD benchmark, we split this dataset into 3 sets: (i) train, (ii) validation and (iii) test.
For each of the $26$ inlier conditions, we split the samples approximately in a $80\%-10\%-10\%$ split for the train-validation-test respectively. 
For the outlier conditions we also split the samples into train, validation and test such that outlier conditions assigned to each of these splits have mutually exclusive conditions as illustrated in Fig.~\ref{fig:graphAbs}. The full list of inlier and outlier conditions are detailed in Table~\ref{tab:list_of_cond}.
The splitting process satisfies the following desiderata:
\begin{itemize}
    \item Patients do not overlap between the splits.
    \item Outlier conditions assigned across the splits are mutually exclusive. 
    \item The number of outlier samples is similar across splits.
    \item The number of outlier conditions is similar across splits, ensuring that the outlier heterogeneity is similar across splits.
    \item Each outlier condition is associated with a worst-case risk level of clinical complications if left untreated (low, medium or high). We ensured a similar distribution of risk categories across the splits to enable downstream analyses on how the OOD detection methods perform across different risk levels.
    \item The distribution of cases with different Fitzpatrick skin types across the splits is similar to enable downstream analysis on how the OOD detection methods perform across different skin types.
\end{itemize}

These requirements can be generalized to establish a reliable benchmark for evaluating OOD methods for any dataset with a long-tailed distribution of classes. The statistics of the resulting splits for our work are detailed in Table~\ref{tab:split_details}.
Note, real-world deployment may involve test outlier classes that were previously seen in training. Thus our more difficult setting of mutually exclusive conditions may underestimate the real-world performance of a OOD detection method. 

\begin{table}[ht]
\centering
\caption{Detailed statistics of our dataset splits. 
The total set of conditions $\SetOfClasses$ is split into inlier $\SetOfInlierClasses$ and outlier conditions $\SetOfOutlierClasses$ depending on whether the condition has sufficient sample size $n_\text{min}$. 
Consequently, the samples $(\obs_i, \cls_i), i=1 \dots N$ in the dataset $\dataset$ are partitioned into $\datasetinlier$ and $\datasetoutlier$ depending on whether $\cls_i \in \SetOfInlierClasses$ or not.  
The inlier dataset $\datasetinlier$ is further split into $\datasettraininlier \cup \datasetvalinlier \cup \datasettestinlier$ sets with approximately 80\% - 10\% - 10\% samples size proportion. 
The outlier dataset $\datasetoutlier$ is split into $\datasettrainoutlier$, $\datasetvaloutlier$ and $\datasettestoutlier$ such that outlier conditions assigned to each of these splits 
are mutually exclusive as illustrated in Fig.~\ref{fig:graphAbs}.
The train dataset is $\datasetTrain = \datasettraininlier \cup \datasettrainoutlier$, validation dataset is $\datasetVal = \datasetvalinlier \cup \datasetvaloutlier$ and test dataset is $\datasetTest = \datasettestinlier \cup \datasettestoutlier$.
We train a deep neural network using $\datasetTrain$ and select our hyper-parameters using $\datasetVal$. The final task is to correctly distinguish $\datasettestoutlier$ from $\datasettestinlier$. }
\vspace{0.5em}
\resizebox{\columnwidth}{!}{
\begin{tabular}{@{}lcccccc@{}}
\toprule
\multicolumn{1}{l}{\multirow{2}{*}{}} & \multicolumn{2}{c}{Train~$\datasetTrain$} & \multicolumn{2}{c}{Validation~$\datasetVal$} & \multicolumn{2}{c}{Test~$\datasetTest$} \\ 
\cmidrule(l){2-3}
\cmidrule(l){4-5}
\cmidrule(l){6-7}
\multicolumn{1}{l}{} & \multicolumn{1}{l}{Inlier} & Outlier & \multicolumn{1}{l}{Inlier} & Outlier & \multicolumn{1}{l}{Inlier} & \multicolumn{1}{l}{Outlier} \\ 
\multicolumn{1}{l}{} & \multicolumn{1}{l}{$\datasettraininlier$} & $\datasettrainoutlier$ & \multicolumn{1}{l}{$\datasetvalinlier$} & $\datasetvaloutlier$ & \multicolumn{1}{l}{$\datasettestinlier$} & \multicolumn{1}{l}{$\datasettestoutlier$} \\
\midrule
 Num. classes & 26 & 68 & 26 & 66 & 26 & 65 \\
 Num. samples & 8854 & 1111 & 1251 & 1082 & 1192 & 937 \\ 
\bottomrule
\end{tabular}
}
\label{tab:split_details}
\end{table}

\subsection{Problem Formulation}
\label{sec:prob_form}
In this section, we provide a more formal definition of our problem statement, introducing the notations that will be used later in the paper. 
We denote the set of labels for all dermatological conditions in our dataset (such as `Eczema', `Acne' etc.; see Fig.~\ref{fig:graphAbs}) by $\SetOfClasses$. Total number of classes is denoted by $C = |\SetOfClasses|$.
Let our long-tailed labeled dataset $\dataset$ with $N$ cases be represented as
$\dataset = \{ (\obs_1, \cls_1), \dots (\obs_N, \cls_N) \}$. Here $\obs_i$ for $i = 1\dots N$ indicates the $i^{\text{th}}$ sample input which corresponds to a set of $j$ image instances $\{I_1, \dots, I_j \}_i$ with $1 \leq j \leq 6$ and $\cls_i$ indicates its corresponding ground truth label with  $\cls_i \in \SetOfClasses$. 
Let $n_{\min}$ indicate the required minimum sample size per condition for any condition $\cls \in \SetOfClasses$ to be considered as inlier. We fix $n_{\min}=100$ similar to~\cite{liu2020deep} for all our experiments.
Given $n_{\min}=100$, we split $\SetOfClasses$ into two mutually exclusive inlier and outlier label sets as $\SetOfClasses = \SetOfInlierClasses \cup \SetOfOutlierClasses$, where we formally define $\SetOfOutlierClasses = \{\cls: \sum_{(\obs', \cls') \in \dataset }  \mathbf{1}(\cls' = \cls)\} < n_{\min} \} $, essentially counting the number of instances that have the label $\cls$ in the dataset. $\mathbf{1}(\cdot)$ is an indicator function.
Consequently, $\SetOfInlierClasses = \SetOfClasses \setminus \SetOfOutlierClasses $.
This also partitions the dataset $\mathcal{D}$ naturally into two mutually disjoint sets $\dataset = \datasetinlier \cup \datasetoutlier$, as an inlier and outlier dataset respectively.

We further split $\datasetinlier$ into $\datasettraininlier$,  $\datasetvalinlier$ and  $\datasettestinlier$. Similarly, we split $\datasetoutlier$ into $\datasettrainoutlier$, $\datasetvaloutlier$ and $\datasettestoutlier$ such that outlier conditions assigned to each of these splits are mutually exclusive as illustrated in Fig.~\ref{fig:graphAbs} and detailed in Sec.~\ref{subsec:data_split}. We denote these mutually exclusive set of outlier conditions by $\SetOfTrainOutlier$, $\SetOfValOutlier$ and $\SetOfTestOutlier$ respectively. The train dataset is $\datasetTrain = \datasettraininlier \cup \datasettrainoutlier$, validation dataset is $\datasetVal = \datasetvalinlier \cup \datasetvaloutlier$ and test dataset is $\datasetTest = \datasettestinlier \cup \datasettestoutlier$.
We train a deep neural network using $\datasetTrain$ and select our hyper-parameters using $\datasetVal$. The final task is to correctly distinguish $\datasettestoutlier$ from $\datasettestinlier$. 

\begin{figure*}[t]
\center{\includegraphics[width=.85\textwidth]{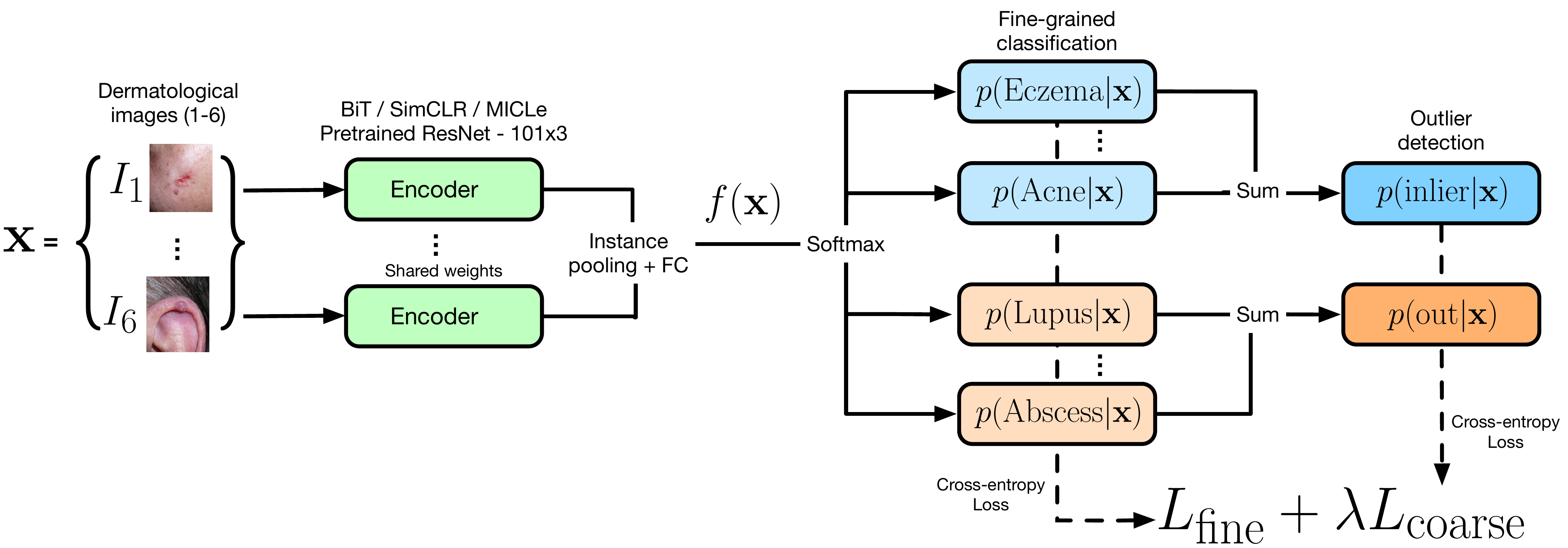}}
\caption{A schematic description of our model architecture and the hierarchical outlier detection (HOD) loss. The encoder (indicated in green) represents the wide ResNet 101x3 model pre-trained with different representation learning approaches (ImageNet, BiT, SimCLR, MICLe). This extracts a feature representation for each of the image instances $\{I_1, \dots, I_6 \}$ of an input case $\obs$. These features are average pooled for all instances and passed through a fully-connected (FC) layer to generate $f(\obs)$ with $512$ dimensions. This output is sent to the HOD loss where a fine-grained and a coarse prediction for inlier (indicated in blue) and training outlier (indicated in orange) are obtained. The coarse predictions are obtained by summing over the fine-grained probabilities as indicated in the figure. The loss is computed by combining them as shown. Note that during test the fine-grained inlier classes and $p(\text{out}|\obs)$ are used.}
\label{fig:hsm}
\end{figure*}

\section{Methods}
\label{sec:methods}
In this section, we describe the model architecture, the proposed hierarchical outlier detection loss and finally the various representation learning and model ensembling strategies we have investigated to boost OOD detection performance.

\subsection{Architectural Details}
\label{subsec:arch}
We illustrate the overall setup in Fig.~\ref{fig:hsm}. As detailed in the Sec.~\ref{sec:prob_form}, we are dealing with a multi-instance dermatology image classification task. Specifically, each case contains $1-6$ image instances, and each case is assigned a primary diagnosis.
We use a common encoder to process all the instances. A wide ResNet $101\times3$ model is chosen as the encoder which is pre-trained with different representation learning approaches detailed later in Sec.~\ref{subsec:rl}.
The encoder provides a feature representation for each image instance in a case. This output is then passed through an instance level average pooling layer, which generates a common feature-map for all the instances for a given case. 
This is passed to a classification head which has an intermediate fully-connected hidden layer with $512$-dimensions followed by a final classification with a softmax layer which provide probabilities for each of the fine-grained inlier and train outlier classes.  
Note, we used ResNet-$101\times3$ as the encoder architecture throughout for simplicity, though extensions to other architectures are straightforward.

\subsection{Hierarchical Outlier Detection Loss}
\label{subsec:hsm}
We build on the recent work by~\cite{thulasidasan2021a} which proposed to have a dedicated abstention class, also referred to as a reject bucket for detecting OOD samples, trained using outlier data. This was shown to be effective in comparison to traditional entropy normalization-based outlier exposure method~\citep{hendrycks2018deep}. 
The outlier samples used for training can have high variability in terms of semantics, acquisition source, resolution etc. 
Encapsulating such a heterogeneous outlier set within a single abstention class can be challenging. One natural mitigation strategy here is to assign multiple abstention classes as possible outputs, representing each of the individual outlier classes available at training via a fine-grained setup.
In the dermatological setting, the approach of using multiple outlier classes is also practical as most often we have access to some labels associated with the training outlier data. Our setup is similar to this; we know the unique outlier classes within our training dataset even though the number of samples of some outlier classes can be as low as $1-2$ samples each. We conjecture that allowing multiple abstention classes has two main advantages: (i) it drastically reduces the burden of fitting a single highly heterogeneous class, hereby allowing a more structured decision boundary, (ii) it provides high capacity to properly model both the inlier and the outlier classes, resulting eventually in richer feature representations. 

Clearly, with a better model for the training outliers, we ultimately want our network to generalize to test outlier samples that are not part of the fine-grained training outlier classes. This can be achieved by encouraging the model to learn features for a high-level semantic separation for the inlier \vs{} outlier classes.
This can potentially assist in modeling the inlier samples by restricting the probability mass within the inlier classes rather than sharing with multiple outlier abstention classes and vice-versa.

Towards this end, we propose employing a Hierarchical Outlier Detection (HOD) loss. This is composed of a fine-grained low level loss and a coarse-grained high level loss.
Given an input sample $\obs$ and its associated label $\cls$ from $\datasetTrain$, let $f(\obs)$ denote the output of the penultimate layer of our model. In our case $f(\obs)$ is a vector of $512$ dimensions.
The predictive probability for a class $c$ is expressed as 
\begin{equation}
    p( c |\obs) = \frac{\exp(\mathbf{w}_c^T f(\obs)+b_c)}{\sum_{c' \in \SetOfInlierClasses \cup \SetOfTrainOutlier} \exp(\mathbf{w}_{c\prime}^T f(\obs)+b_{c'})},
\end{equation}
where $\mathbf{w}_c$ and $b_c$ are the weights and bias of the last layer for class $c$, and $ \sum_{c \in \SetOfInlierClasses \cup \SetOfTrainOutlier } p( c |\obs) = 1$. 
Note that the network provides an output of dimension $|\SetOfInlierClasses \cup \SetOfTrainOutlier|$, with $|\SetOfTrainOutlier|$ abstention classes.

The two level hierarchical loss is then constructed based on the predictive probabilities. 
The fine-grained loss is defined as the cross entropy loss, \ie{} the negative log-likelihood, of the true label class $y$,
\begin{equation}
L_{\text{fine}} = - \sum_{c \in \SetOfInlierClasses \cup \SetOfTrainOutlier} \mathbf{1}(\cls=c) \log p(c |\obs).
\end{equation}
For the coarse-grained loss, we first define the probability of $\mathbf{x}$ being an inlier 
as the sum of the probabilities of the fine-grained inlier classes in $\SetOfInlierClasses$, $p(\text{inlier}|\mathbf{x}) = \sum_{c \in \SetOfInlierClasses} p(c | \mathbf{x})$ . Similarly the probability of being an outlier 
is the sum of the probabilities of the fine-grained outlier classes in $\SetOfTrainOutlier$, $p(\text{out}| \mathbf{x}) = \sum_{c \in \SetOfTrainOutlier} p(c | \mathbf{x})$.
This is shown in Fig.~\ref{fig:hsm}.
The coarse-grain level is a binary classification between classes $\{\text{inlier}, \text{out}\}$. 
The coarse-grain loss is then defined as 

\begin{equation}
L_{\text{coarse}} = - \sum_{c \in \{\text{inlier}, \text{out}\} } \mathbf{1}(\cls=c) \log p(c |\obs).
\end{equation}

As a training objective, we propose directly optimizing the following combined overall loss of
\begin{equation}
    L_{\text{overall}} = L_{\text{fine}} + \lambda L_{\text{coarse}},
\end{equation}
where the hyper-parameter $\lambda$ dictates the relative importance of $L_{\text{coarse}}$ in comparison to $L_{\text{fine}}$. Fig.~\ref{fig:hsm} illustrates the hierarchical loss structure.
We define the OOD score $\mathcal{U}(\obs)$ and confidence scores $\mathcal{C}(\obs)$ as 
\begin{align}
    \mathcal{U}(\obs) &= p(\text{out}|\obs), \\
    \mathcal{C}(\obs) &= 1-p(\text{out}|\obs).
\end{align}

At test time, the individual probabilities for the $|\SetOfTrainOutlier|$ abstention classes do not carry any importance because none of them show up in the test set. The sum of these probabilities indicating the OOD score $\mathcal{U}(\obs)$ is used.

We argue that the HOD loss provides two major benefits for OOD detection. First, the fine-grained loss is helpful for reducing OOD data heterogeneity per abstention class, hereby allowing more structured decision boundaries. Second, the coarse-grained loss plays a dynamic label smoothing role, improving the generalization for unseen outlier detection.
In particular, for an OOD input $(\obs, \cls)$ where $\cls=c, c \in \SetOfTrainOutlier$, the log-likelihood it contributes to is,
\begin{equation}
    \log p(c|\obs) + \lambda \log \left( p(c|\obs)+ \sum_{c' \in \SetOfTrainOutlier \setminus \{c\}} p(c'|\obs) \right),
\end{equation} 
where the first term is for the fine-grained loss and the second is for the coarse-grained loss.
To minimize the loss, equivalently maximizing the log-likelihood, the model can either through gradient descent increase $p(c|\obs)$, or increase the $p(c'|\obs), c' \in \SetOfTrainOutlier \setminus \{c\}$ for the rest of the OOD classes. 
In other words, the fine-grained loss is to increase the probability for the particular class, and the coarse-grained loss is to increase the probability mass for the rest of the OOD classes. 
Even though that data point is for the OOD class $c$, the parameters relevant to the rest of the OOD classes $c'$ are also updated with respect to the gradient of this loss, leading to a similar effect as label smoothing. 
This property is extremely helpful when the test OOD classes are mutually exclusive from the training OOD classes. 
While the model is trained using training OOD classes, the test OOD data will not have high probability for any specific training OOD classes $p(c|\obs), c \in \SetOfTrainOutlier$, as the test OOD class is unseen. But because of the coarse-grained loss, it may have relatively high probability in the aggregated probability for OOD classes $p(\text{out}|\obs)$, resulting in to a better generalization for OOD performance. 

One interpretation of this combined loss is to think along the lines of data augmentation. Let us imagine that each input $\obs$ is associated with a coarse grained label and a fine-grained label. We duplicate this sample $a$ times and pair it with its fine-grained labels. Similarly, we duplicate the same sample $b$ times and pair them with its coarse labels. The combined log-likelihood contribution of the sample $\obs$ augmented $a+b$ times would be equivalent to our loss function $L_{\text{fine}} + \lambda L_{\text{coarse}}$, where $\lambda = b/a$.

Note that \citet{yan2015hd} also incorporate the hierarchy among image classes into the model but in a different way: independent fine grained classification heads are constructed for each high level class. In addition, their study is for improving the inlier prediction accuracy, not for OOD detection. To the best of our knowledge, we are the first to use such a hierarchical loss for OOD detection.

\subsection{Representation Learning}
\label{subsec:rl}
Representation learning has been proven effective for semi-supervised learning where the amount of training data is limited~\citep{kolesnikov2019big, chen2020simple}. While the main goal of representation learning is enabling label efficiency in downstream tasks, in this article, we will mainly explore the effectiveness of representation learning for downstream OOD detection task. 
We hypothesize that generic features may be learned during the pre-training phase. While these features may be discarded when training using only a supervised objective as they may not be directly useful for classification, we may be able to leverage these for OOD detection.

In image classification, a common representation learning approach is via a model pre-trained on ImageNet. We use a wide ResNet-$101\times3$ feature extractor pre-trained on natural images from the ImageNet dataset as the baseline representation learning approach. An ImageNet based pre-training strategy was also used in~\citet{liu2020deep}.

In addition, we explore two other representation approaches for our task: Big Transfer (BiT) as a supervised approach, and contrastive training as a self-supervised approach. These are detailed below.

\subsubsection{Big Transfer (BiT) using JFT}
Big Transfer (BiT) is a large scale supervised pre-training for visual representation learning, introduced by~\citet{kolesnikov2019big}. This was later extended to different medical applications and demonstrated promising results~\citep{mustafa2021supervised}. The method provides a recipe for efficient transfer learning with minimal hyper-parameter tuning and its effectiveness is demonstrated in a plethora of vision tasks. The main architectural changes include: (i) using Group Normalization~\citep{wu2018group} instead of Batch Normalization and (ii) inclusion of weight standardization~\citep{qiao2019weight}, which aids effective transfer learning with variable batch size.
In this paper, we use a wide ResNet-$101\times3$ BiT model pre-trained on the large JFT dataset~\citep{sun2017revisiting}. This is referred as BiT-L. The purpose of this experimental choice is to investigate if adding considerably more natural images helps learn generic features to improve OOD detection compared to the ImageNet baseline.

\subsubsection{Contrastive Training}
We use two contrastive self-supervised learning methods for our application, which has been shown to be highly effective at representation learning~\citep{chen2020simple}. 
In contrast to ImageNet and BiT-L pre-training, we leverage unlabeled data from the target dermatology domain for pre-training our encoder optimizing the contrastive objective.

First, we use SimCLR~\citep{chen2020simple} based contrastive pre-training due to its simplicity.
SimCLR based pre-training has already been demonstrated to be effective on benchmark OOD tasks~\citep{winkens2020contrastive} potentially due to the rich representations learnt. For our application we use SimCLR to pre-train a wide ResNet-$101\times3$ model. This model was first pre-trained on ImageNet and then further trained on a set of unlabeled dermatology images using the contrastive objective. This includes images from both $\datasetTrain$ and from additional unlabeled dermatology dataset detailed in Sec.~\ref{sec:dataset_description}. The data augmentations used in this contrastive pre-training are random color augmentation, random crops ($224\times224$ pixels), Gaussian blur and random flips.
Our hope is that using dermatological images for pre-training will aid learning of domain specific representations for downstream OOD detection. 

Second, as our downstream classification task is multi-instanced in nature, we also use a multi-instanced version of contrastive training termed MICLe~\citep{azizi2021big}.
In contrast to SimCLR which tries to minimize the distance between two augmented versions of the same image, MICLe aims at minimizing the distance between two augmented image instances of the same case in the feature space. This modification help to learn representations that can distinguish each multi-instanced case from each other.
The same unlabeled dermatology dataset as SimCLR was used for training.

We compare the OOD performance using ImageNet, BiT, SimCLR and MICLe representation learning, with HOD loss and with single reject bucket in the results.

\subsection{Representational Diversity in Ensembles}
\label{sec:ensemble}
We ensemble multiple models to further improve the OOD detection performance. 
For a set of $T$ ensemble members trained with random initialization and random shuffling of training inputs, we use the average of the predicted OOD score $\bar{p}(\text{out}|\obs)=\frac{1}{T} \sum_{t=1}^T p_t(\text{out}|\obs)$, as the final OOD score. Here $p_t(\text{out}|\obs)$ indicates the OOD score for the $t^{\text{th}}$ model.
If the model uses HOD loss, the OOD score is the sum of the probabilities of the fine-grained outlier classes in $\SetOfTrainOutlier$. If the model uses reject bucket loss, the OOD score is the probability of the reject bucket. 
For each of the proposed models, we choose $T=5$ because of diminishing returns beyond that~\citep{lakshminarayanan2016simple}.

Note that performance boost in deep ensembles depends on the model diversity introduced by random initialization of the network weights. In our setup, as we are using pre-trained models, random initialization is done only for the weights of the last layer. This reduces the diversity.
To introduce more diversity, we also ensemble models trained with different representation learning and objective function to increase the diversity of the ensemble members. 
Given a set of candidate models, we use the greedy search algorithm proposed by~\citet{wenzel2020hyperparameter} to select the best subset of models whose ensemble gives the best OOD performance.
In particular, we first collect all models we have trained using different representation learning and objectives: ImageNet initialized models with and without HOD loss, BiT models with and without HOD loss, SimCLR models with and without HOD loss and MICLe models with and without HOD loss. 
We greedily grow an ensemble, until reaching a fixed ensemble size, by selecting with replacement the model leading to the best improvement of OOD performance on the validation dataset.
We refer to the selected ensemble as the diverse ensemble. 
We believe this diversity will boost OOD performance leveraging the complementary features learnt from the diverse representation learning approaches and objective functions we investigate. 
We compare the performance of this diverse ensemble with vanilla ensembles in the results.

\section{Experimental Results}
\label{sec:exp_res}
In this section, we present the experimental setup and the results of our experiments which includes ablation of the different components we propose and comparison to baseline approaches. 

\begin{table*}[t]
\centering
\caption{Comparison with existing methods and perform ablations for Hierarchical Outlier Detection (HOD) loss components. We use a BiT-L pre-trained ResNet-101x3 model as the backbone. For each method we trained $5$ models with random last layer initialization and batch shuffling, and reported mean $\pm$ standard deviation for each score. We report inlier accuracy of the model along with OOD detection metrics. Note that the first two methods indicated by $\star$ do not use outlier data in the training stage. The best result in each column is indicated in \textbf{bold}.}
\vspace{0.5em}
\begin{tabular}{lcccc}
 \toprule
 \multirow{2}{*}{Method}
  & \multicolumn{3}{c}{OOD detection metrics} &\\
\cmidrule{2-4}
&  AUROC ($\uparrow$) & FPR @ 0.95 TPR ($\downarrow$) & AUPR-in ($\uparrow$) & Inlier accuracy ($\uparrow$)\\
 \otoprule
 BiT-L + MSP$^\star$ & $72.1 \pm 0.2$ & $81.3 \pm 1.0$ & $76.5 \pm 0.1$ & $69.1 \pm 0.4$ \\
 BiT-L + Mahalanobis$^\star$ & $68.1 \pm 0.3$ & $90.4 \pm 0.5$ & $75.2 \pm 0.3$ & $69.1 \pm 0.4$ \\
 \midrule
 BiT-L + Outlier exposure + MSP & $72.9 \pm 0.3$ & $81.0 \pm 0.7$ & $77.4 \pm 0.5$ & $68.2 \pm 0.4$ \\
 BiT-L with reject bucket      & $75.6 \pm 0.9$          & $72.3 \pm 1.7$          & $78.3 \pm 0.9$ &  $72.8 \pm 0.3$  \\
 \midrule
 BiT-L with fine-grained outlier ($\lambda = 0$)  & $78.3 \pm 0.4$          & $72.9 \pm 0.3$ & $81.3 \pm 0.5$& $\mathbf{75.6} \pm 0.3$ \\
 BiT-L + HOD ($\lambda = 0.1$) & $\mathbf{79.4} \pm 1.0$ & $\mathbf{65.9} \pm 1.1$ & $\mathbf{81.8} \pm 1.1$   & $74.0 \pm 0.6$ \\
 BiT-L + HOD ($\lambda = 0.5$) & $75.1 \pm 0.8$          & $76.3 \pm 2.6$          & $78.8 \pm 0.8$  & $62.6 \pm 0.7$ \\
 BiT-L + HOD ($\lambda = 1$)    & $71.3 \pm 1.1$          & $80.7 \pm 2.2$          & $75.2 \pm 1.0$   & $50.5 \pm 1.3$\\
 \bottomrule
\end{tabular}
\label{tab:hsm_ablation}
\end{table*}

\subsection{Experimental Settings and Evaluation Metrics}
Here we detail the experimental setting for all the models and introduce the evaluation metrics we use for comparing different methods.
\paragraph{Experimental settings}
We use the train split $\datasetTrain$ (as detailed in Table~\ref{tab:split_details}) for training all the models. We use data augmentation for training. This includes: random horizontal and vertical flips, random variations of brightness (max intensity = $0.1$), contrast (intensity = [$0.8-1.2$]), saturation (intensity = [$0.8-1.2$]) and hue (max intensity = $0.02$), random Gaussian blurring using standard deviation between $0.01$ and $7.0$ and random rotations between $-150^{\circ}$ and $150^{\circ}$.
We use a Stochastic Gradient Descent optimizer with momentum with exponentially decaying learning rate for training. Each model is trained for $10,000$ steps. Convergence of all the models were ensured on the validation set. We use a batch-size of $16$ cases for training.
We use the validation split $\datasetVal$ to set the hyper-parameters (\eg{} initial learning rate, decay factor, momentum) and to select the best checkpoints for all the models. Checkpoint selection was based on the OOD Area under Receiver Operating Characteristic Curve (AUROC).
AUROC provides a quantitative measure of predictive performance for
outliers using the OOD score, with higher values indicating higher
power in discriminating inliers and outliers using the OOD score.
We report all the results in the following sections on all the samples in the held out test split $\datasetTest$. Note that all the metrics used for evaluation do not require selection of a fixed operating point.

\paragraph{Evaluation metrics} 
For evaluating the OOD performance for different methods, we use 3 commonly used metrics: (i) AUROC (higher is better), (ii) False positive rate at $95\%$ true positive rate (FPR $@~95\%$ TPR, lower is better) and (iii) Area under inlier precision-recall curve (AUPR-in, higher is better). 
Along with the OOD metrics, we also track the inlier accuracy of the models to investigate any possible trade-off between accuracy and OOD performance of the models. The inlier accuracy is computed only on the test inlier set, comparing the ground-truth to the top-1 inlier prediction.

\subsection{Comparison of HOD Loss with Existing Methods}
In this section we compare our proposed HOD loss with existing methods and ablation of the different parts of the HOD loss. As an architectural backbone for this ablation we choose the BiT-L initialized wide ResNet $101\times3$. 
First we investigate the most commonly used OOD detection methods: MSP~\citep{hendrycks2016baseline} and Mahalanobis distance~\citep{lee2018simple}. Note that these baselines do not use outlier samples in the training process. We present the results in the first two rows of Table~\ref{tab:hsm_ablation}. We observe that MSP baseline outperforms Mahalanobis baseline by $4$ AUROC points.
Next, we investigate existing methods which use training outliers: outlier exposure (OE) with MSP~\citep{hendrycks2018deep} and reject bucket~\citep{thulasidasan2021a}. We present the results in Table~\ref{tab:hsm_ablation}. 
We observe that OE with MSP outperforms the MSP baseline by $0.8$ AUROC points.
The reject bucket baseline further outperforms OE by $2.7$ AUROC points.
It is clear from the results that methods using outliers during training are better than the methods not using them.
Among all the existing methods the reject bucket method has the best OOD detection performance and we use it for comparison for the next experiments.

Note that the reject bucket based baseline uses a single abstention class to encapsulate the OOD data~\citep{thulasidasan2021a}. 
As illustrated in Sec.~\ref{subsec:hsm} HOD introduces two modifications on top of this: (i) expanding the reject bucket into fine-grained training outlier classes, and (ii) inclusion of a coarse loss term along with fine-grained loss. 
We also explored different choices of $\lambda$ which dictates the relative contribution of fine-grained and coarse-grained loss terms. We present the results for all the settings on the test set in Tab.~\ref{tab:hsm_ablation}. 

\begin{table*}[h]
\centering
\caption{Comparison of HOD \vs{} reject bucket for different representation learning approaches. We report inlier accuracy of the model along with OOD detection metrics. For each model we trained $5$ models with different last layer initialization and batch shuffling, and reported mean $\pm$ standard deviation for each score. For all the HOD models we use $\lambda = 0.1$.}
\vspace{0.5em}
\begin{tabular}{lcccc}
 \toprule
 \multirow{2}{*}{Method}
   & \multicolumn{3}{c}{OOD detection metrics}& \\
\cmidrule{2-4}
 & AUROC ($\uparrow$) & FPR @ 0.95 TPR ($\downarrow$) & AUPR-in ($\uparrow$) & Inlier accuracy ($\uparrow$) \\
 \otoprule
 ImageNet + reject bucket    & $74.7 \pm 0.5$ & $78.9 \pm 0.7$ & $78.1 \pm 0.8$  & $72.0 \pm 0.6$ \\
 ImageNet + HOD              & $77.0 \pm 0.9$ & $70.3 \pm 3.2$ & $79.4 \pm 0.7$  & $69.7 \pm 1.1$ \\
 \midrule
 BiT-L + reject bucket     & $75.6 \pm 0.9$ & $72.3 \pm 1.7$ & $78.3 \pm 0.9$  & $72.8 \pm 0.3$\\
 BiT-L + HOD               & $79.4 \pm 1.0$ & $65.9 \pm 1.1$ & $81.8 \pm 1.1$  & $74.0 \pm 0.6$  \\
 \midrule
 SimCLR + reject bucket      & $75.2 \pm 1.0$ & $77.4 \pm 1.8$ & $78.1 \pm 0.8$  & $74.3 \pm 0.3$ \\
 SimCLR + HOD                & $77.2 \pm 0.9$ & $71.3 \pm 2.7$ & $79.8 \pm 0.6$  & $69.0 \pm 1.5$\\
 \midrule
 MICLe + reject bucket       & $76.7 \pm 0.7$ & $72.3 \pm 2.3$ & $79.4 \pm 0.9$  & $74.4 \pm 0.9$\\
 MICLe + HOD                 & $78.8 \pm 0.6$ & $68.9 \pm 1.2$ & $81.3 \pm 0.9$  & $70.7 \pm 1.3$ \\
 \bottomrule
\end{tabular}
\label{tab:backbone_ablation}
\end{table*}

First we observe that expanding the abstention class to fine-grained outlier classes improves OOD detection AUROC by $3$ points and inlier accuracy by $3$ points compared to reject bucket baseline. This indicates that assigning multiple class-specific buckets to outlier samples is better than encapsulating all highly heterogeneous outlier samples in a single abstention class both for OOD detection and inlier accuracy. 
Inclusion of a coarse loss with $\lambda=0.1$ further boosts performance by $1$ point AUROC with a slight decrease of $1.6$ points accuracy compared to having multiple abstention classes \ie{} $\lambda=0$. The inclusion of the coarse loss had a large impact in reducing FPR $@~95\%$ TPR by $7$ points which was not observed with $\lambda=0$. 

Also, we observe that assigning higher weights to the coarse loss $\lambda \in \{ 0.5, 1.0 \}$ drastically degrades both the inlier accuracy and OOD detection performance. We believe this is due to the strong label smoothing regularizing effect that the coarse loss provides. 
Setting a very high $\lambda$ is similar to training a model for a binary inlier/outlier classification only. Note that inlier accuracy drops drastically with higher values of $\lambda$. 
Thus, we fix the value of $\lambda$ to $0.1$ and use this setting for the following experiments. 

Note that Mahalanobis method~\citep{lee2018simple} can also be used as an OOD score. This involves fitting class conditional Gaussians for the inliers in the high-dimensional feature space and use the Mahalanobis distance for a test sample to the fitted distribution as an OOD score. We performed experiments using the $512$ dimensional $f(\obs)$. We observed that it consistently performed poorly compared to using $p(\text{out}|\obs)$ as an OOD score. For BiT-L with reject bucket, on the validation set $p(\text{out}|\obs)$ had an AUROC of $77\%$, whereas the Mahalanobis method had an AUROC of $70\%$. We believe this drop is due to the under-fitting of the Gaussians in the feature space. Fitting a Gaussian in a $512$ dimensional space requires estimating $512 + 512\times512$ parameters for the mean vector and co-variance matrix. 
Note that for our application, we also have a class imbalance among the inlier conditions (see Fig.~\ref{fig:graphAbs}).
The sample count for some inlier classes is not sufficient for reliably fitting a high-dimensional Gaussian in the feature space.
The Mahalanobis method works well in most public benchmarks as they have a balanced inlier class distribution.
Thus, we only use $p(\text{out}|\obs)$ as the OOD score in this work.

\subsection{Effect of Different Representation Learning Methods}
\label{sec:ablation_rl}
In this section we investigate how different types of representation learning (see Sec.~\ref{sec:methods}) helps in OOD detection in conjunction with HOD: ImageNet pre-trained, BiT-L pre-trained, SimCLR pre-trained and MICLe pre-trained models. 
First, as shown in Table~\ref{tab:backbone_ablation}, we observe that including HOD loss improves OOD performance compared to reject bucket for all the four representation learning methods. The AUROC increases by $2.3$ points, $3.8$ points, $1.1$ points and $2.1$ points for ImageNet, BiT-L, SimCLR and MICLe pre-trained models respectively. The boost is much higher in BiT-L in comparison to others. 
HOD also boosts the inlier accuracy for BiT-L checkpoints by 1.2 points. However, we observed a drop in inlier accuracy by $2.3$, $5.3$ and $3.7$ points with HOD for ImageNet, SimCLR and MICLe respectively. 
Note that after training the model checkpoint was selected based on best AUROC performance on validation split. For ImageNet, SimCLR and MICLe with HOD loss, inlier accuracy and AUROC peaks at different stages of the training and there seems to be a trade-off between the two. This might be a possible reason for this drop in inlier accuracy. 

Comparing the natural image based pre-training methods (ImageNet, BiT-L), we observe that BiT-L has a better inlier accuracy and OOD performance. We can conclude that using much larger-scale datasets (JFT) for pre-training helps not only for better classification performance but also for better OOD detection task.
Comparing the contrastive training based methods (SimCLR, MICLe), we observe that although MICLe and SimCLR yields similar inlier accuracy, OOD performance of MICLe is much better than SimCLR. SimCLR based pre-training learns to distinguish every image and does not consider the multi-instance aspect of our task. By considering the multi-instance aspect, MICLe may have learned case-specific features that are more useful for detecting case-level OOD samples.

Comparing the reject bucket based models, we also observe that contrastive learning based models have higher inlier accuracy in comparison to natural image based representation learning ones. We believe this is due to the additional unlabeled dermatology images used for contrastive learning which aided in learning domain specific features. This played a major role only for models with reject bucket.

\begin{table*}[ht]
\centering
\caption{Comparison of ensembles of different models. We report inlier accuracy of the model along with OOD detection metrics. For each model reported in the table we limit the ensemble size to $5$ models for direct comparison to Table~\ref{tab:backbone_ablation}. The models in the diverse ensemble were selected using a greedy algorithm on validation set. For all the HOD models we use $\lambda = 0.1$. The diverse ensemble is comprised of 3 BiT-L + HOD models and 2 MICLe + HOD models. The best result in each column is indicated in \textbf{bold}.}
\vspace{0.5em}
\begin{tabular}{lcccc}
 \toprule
 \multirow{2}{*}{Method}
   & \multicolumn{3}{c}{OOD detection metrics} & \\
\cmidrule{2-4}
 & AUROC ($\uparrow$) & FPR @ 0.95 TPR ($\downarrow$) & AUPR-in ($\uparrow$) & Inlier accuracy ($\uparrow$) \\
 \otoprule
  ImageNet + reject bucket + Ensemble        & $76.4$ & $75.3$ & $79.9$ & $72.9$\\
  ImageNet + HOD + Ensemble  & $79.2$ & $70.6$ & $81.8$ & $70.9$ \\
  \midrule
  BiT-L + reject bucket + Ensemble         & $77.8$ & $71.0$ & $80.6$ & $73.8$ \\
  BiT-L + HOD + Ensemble   & $81.6$ & $62.6$ & $83.9$ & $75.6$ \\
  \midrule
  SimCLR + reject bucket + Ensemble.         & $77.0$ & $76.4$ & $79.8$  & $75.1$ \\
  SimCLR + HOD + Ensemble    & $78.7$ & $70.4$ & $81.5$  & $71.3$ \\
  \midrule
  MICLe + reject bucket + Ensemble           & $79.0$ & $71.5$ & $82.1$ & $75.8$ \\
  MICLe + HOD + Ensemble     & $80.5$ & $67.5$ & $83.2$ & $72.4$  \\
  \midrule
  Diverse ensemble  & $\mathbf{83.0}$ & $\mathbf{61.4}$ & $\mathbf{85.8}$ & $\mathbf{76.3}$\\
 \bottomrule
\end{tabular}
\label{tab:ensemble_ablation}
\end{table*}

\subsection{Comparison of Ensembling Strategies}
\label{sec:ens_ablation}
In this section, we study different ensembling strategies and investigate their efficacy for OOD detection.
For this study, we use all the different representation learning methods detailed in Sec.~\ref{sec:ablation_rl}, with reject bucket and with HOD loss.
Firstly, we investigate the vanilla ensembling strategy. For each of the methods, we ensemble five models trained independently with random initialization and random data shuffling. 
Note that as we used pretrained models for initialization, only the final fully connected layer and classifier layer weights were randomly initialized.
Table~\ref{tab:ensemble_ablation} shows the performance of each method after ensembling. 
Compared with Table~\ref{tab:backbone_ablation}, it is clear that for every method, ensembling improves both inlier accuracy and OOD detection performance in comparison to their single-model counter-parts. 

Secondly, we investigate ensembling models with different representation learning methods as described in Sec.~\ref{sec:ensemble}.
For the purposes of testing the diverse ensemble strategy, we pool together all the $40$ models for ImageNet, BiT-L, SimCLR and MICLe with reject bucket and with HOD and employ a greedy search algorithm~\citep{wenzel2020hyperparameter} to select a set of $5$ models. 
The selection criteria metric for the greedy search algorithm was the mean of all 3 OOD metrics: AUROC, $1 - $ FPR $@~95\%$ TPR and AUPR-in. The ensemble with the highest validation set performance was selected for use (Table~\ref{tab:ensemble_ablation}). The selected five models are three BiT-L pre-trained models with HOD loss and two MICLe pre-trained models with HOD loss. 
This diverse ensemble achieves the highest AUROC of $83\%$, AUPR of $85.8\%$, inlier accuracy of $76.3\%$, and the lowest FPR $@~95\%$ TPR of $61.4\%$.
The diverse ensemble selected only HOD loss based models, indicating they were stronger candidates than the reject bucket based models. 
Further, note that the greedy search algorithm selects models from from two different representation learning (BiT and MICLe).
We believe natural image based pre-trained BiT-L and contrastive learning based pre-trained MICLe models learn complementary features. Training with natural images may have helped by learning more generic low level features, whereas contrastive learning may have helped in learning more dermatology-specific features during its pre-training phase. 
This promotes learning both high-level and low-level features which might not be very useful for inlier classification, but might be useful for identifying previously unseen OOD examples. 
This complimentary nature may have enhanced the diversity during the selection process.
Furthermore, note that the drop in inlier accuracy performance of the HOD loss in MICLe models for vanilla ensembles is compensated by the diverse ensemble.
We use this diverse ensemble model as our final model for the subsequent analysis.

\section{Discussion}
\label{sec:discuss}

As the final model is determined, in this section we discuss a few factors that may provide useful guidance for further improvement. 
We also perform downstream analysis of the model's performance for different subgroups of risk levels and skin types, and we perform a trust analysis to better understand the model's total clinical implications. 

\begin{table}[ht]
\centering
\caption{Ablation over training outlier heterogeneity. We use the BiT-L model for this. BiT-L + MSP indicates training without any outliers with max-of-softmax probabilities as the OOD score. BiT-L + HOD-xx indicates the models with varying train outlier heterogeneity (xx indicates the number of abstention classes which corresponds to number of outlier classes). The last row indicated by BiT-L + HOD-68 uses the entirety of our training outlier set. For each setting we trained $5$ models and reported mean $\pm$ standard deviation for OOD detection AUROC on the test set. For all the HOD models we use $\lambda = 0.1$. The best result is indicated in \textbf{bold}.}
\vspace{0.5em}
\begin{tabular}{lccc}
 \toprule
  & $\datasettrainoutlier$ &  $\datasettrainoutlier$ & AUROC  \\
  & classes & samples & ($\uparrow$) \\
 \otoprule
 BiT-L + MSP & $0$ & $0$ & $72.1 \pm 0.2$ \\
 \midrule
 BiT-L + HOD-17 & $17$ & $230$ & $73.1 \pm 0.7$ \\
 BiT-L + HOD-34 & $34$ & $483$ & $77.4 \pm 0.7$ \\
 BiT-L + HOD-51 & $51$ & $768$ & $78.7 \pm 1.0$ \\
 BiT-L + HOD-68 & $68$ & $1111$ & $\mathbf{79.4} \pm 1.0$ \\
 \bottomrule
\end{tabular}
\label{tab:num_train_ablation}
\end{table}

\subsection{Available Training Outlier Data}
In this section, we investigate the OOD performance of the model with varying amounts of training outlier data. We hypothesize that both quantity (number of train outlier samples) and quality (number of outlier train classes) play a major role in efficiently detecting outliers during deployment. To investigate this we use different proportions of training outlier data to train models and presented the results in Table~\ref{tab:num_train_ablation}. We use the BiT-L with HOD loss for this experiment. For each setting, we train $5$ different models and report the mean and standard deviation of OOD detection AUROC. We use the same validation set for checkpoint selection and hyper-parameter tuning and reported results on the same test set for all settings. 

As a first setup, we show the results of not using any outliers during training. For this we trained a BiT-L pre-trained model only with inlier samples and used max-of-softmax (MSP) probabilities as the OOD score~\citep{hendrycks2016baseline}. We indicate this as BiT-L + MSP in Table~\ref{tab:num_train_ablation}.
Following that we uniformly increase more training outlier classes and samples. 
Each entry is indicated by BiT-L + HOD-xx, where xx corresponds to number of training outlier class \ie{} number of abstention classes used in training.
The last row includes the entirety of our training outlier set with $68$ outlier classes as indicated in Table~\ref{tab:split_details}. We observe that as we increase the training outlier heterogeneity and quantity the OOD detection performance increases consistently. Given the lack of plateauing, introducing additional training outlier classes and samples may potentially increase OOD detection performance further.

\begin{figure*}[h]
     \centering
     \begin{subfigure}[b]{0.55\textwidth}
         \centering
         \includegraphics[width=\textwidth]{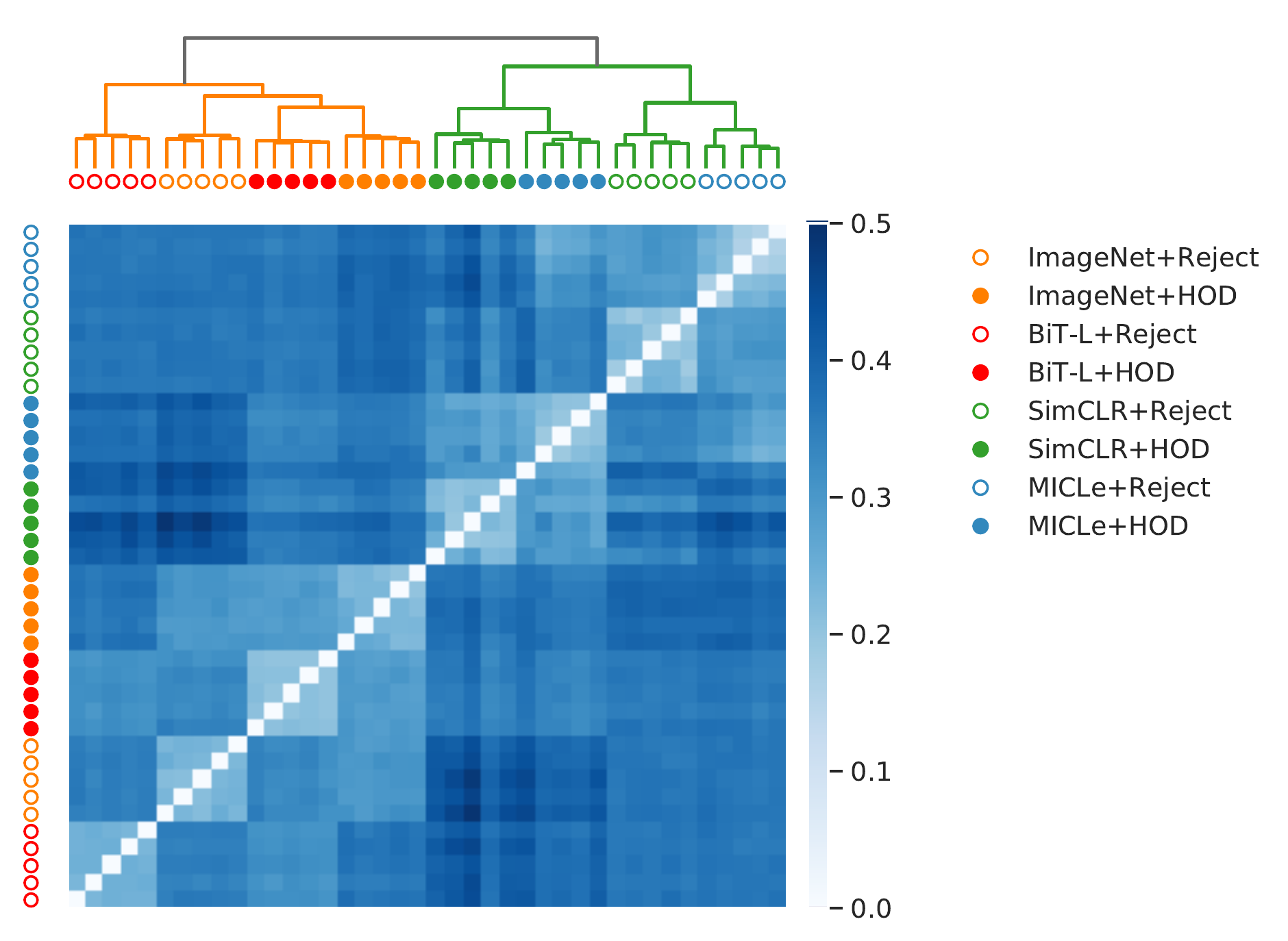}
         \caption{}
         \label{fig:heatmap_div}
     \end{subfigure} 
     \begin{subfigure}[b]{0.38\textwidth}
         \centering
         \includegraphics[width=\textwidth]{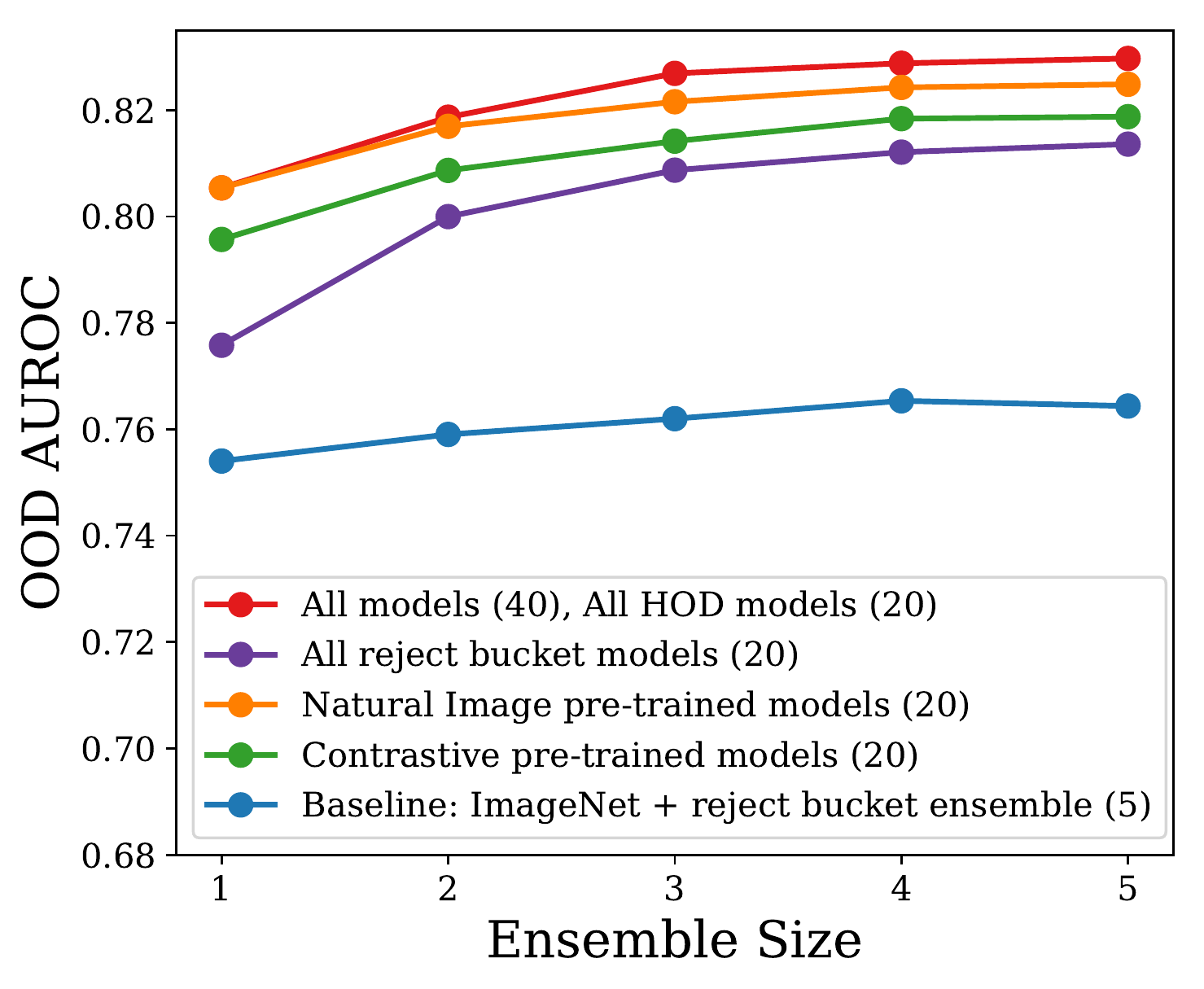}
         \caption{}
         \label{fig:auc_vs_diveristy}
     \end{subfigure}
        \caption{This figure shows the diversity among our pool of $40$ models that are trained using different representation learning methods, different objective functions and different classification layer initialization, and how the diverse ensemble helps to improve OOD detection performance. 
        (a) The heatmap for pairwise model diversity for all the models. The diversity is computed based on Eqn.~\ref{eq:diversity}. Darker shades of blue indicates higher diversity. 
        The models include four representation methods combined with two objective functions \ie{} $8$ different configurations, denoted by different symbols (indicated in legend). For each configuration, $5$ different models with random last layer initialization and random shuffling of the input data batches were trained. This accounts for the overall $40$ models in our pool.
        Natural image pre-trained models \ie{} BiT-L and ImageNet are indicated by red and orange circles respectively.  Contrastive pre-trained models \ie{} SimCLR and MICLe are indicated by green and blue circles respectively. The models trained with the HOD loss are indicated by filled circles, whereas the models trained with the reject bucket loss are indicated by hollow circles.
        We also perform a hierarchical clustering using the diversity as a distance and show the corresponding dendrogram at the top.
        We observe that representation learning and loss function provides the highest diversity to the pool.
        (b) The OOD detection AUROC performance of selected diverse ensemble by greedy search algorithm on different pools of models with varying ensemble size is shown here. 
        The baseline vanilla ensemble (blue line) was based on the $5$ ImageNet pre-trained models with reject bucket loss. The other pools are: $20$ models that are pre-trained using reject bucket loss (purple line), $20$ models that are contrastive pre-trained (green line), $20$ models that are pre-trained using natural images (orange line), $20$ models trained using HOD loss (red line), and all $40$ models (red line). Note that similar models were selected over pool of all models and pool of HOD loss models, thus both indicated together in red. The pool size for each is indicated in the legend.}
        \label{fig:heatmap}
\end{figure*}

\subsection{Diversity in ensembles}
\label{sec:diversity_analysis}

In Sec.~\ref{sec:ens_ablation}, we demonstrated that diverse ensemble selected by a greedy search algorithm outperforms all the vanilla ensemble models. In this section, we investigate how diversity plays a role in achieving this performance boost.
In our pool of $40$ models we have diversity in the following aspects: (i) representational diversity (pre-training strategies with ImageNet, BiT-L, SimCLR and MICLe), (ii) diversity in objective function (two different losses: reject bucket loss and HOD loss) and (iii) diversity due to last layer initialization and random input shuffling (vanilla ensembling).

To quantify diversity between any two models $M_a$ and $M_b$, we use the average predictive disagreement between them similar to~\citep{fort2019deep}. For each input sample $\obs_i \in \datasetTest$, let the top-1 predicted class given by $M_a$ be $y_i^{M_a}$ and by $M_b$ be $y_i^{M_b}$. The diversity is given as

\begin{equation}
D(M_a, M_b) = 1 - \mathbb{E}_i [y_i^{M_a} == y_i^{M_b}],
\label{eq:diversity}
\end{equation}
\noindent
where $\mathbb{E}[\cdot]$ indicates the expectation operator. Note that for computing top-1 prediction we perform an $\argmax$ over the output class probabilities, which includes all the fine-grained inlier conditions and a single outlier class given by $p(\text{out}|\obs)$.

In Fig.~\ref{fig:heatmap_div}, we present a $40\times40$ diversity heatmap for all the models in our pool. The darker shades indicate higher diversity and the lighter shades indicates low diversity.
Using the diversity in Eqn.~\ref{eq:diversity} as the pair-wise distance between models, we apply hierarchical clustering using Ward's minimum variance, over the pool of models and show the corresponding dendrogram at the top of Fig.~\ref{fig:heatmap_div}.
At the highest level of the dendrogram, we observe a separation between all contrastive pre-trained (SimCLR, MICLe) models and natural image pre-trained (ImageNet, BiT-L) models. This indicates that different representation learning provides the most diversity within our pool.
At the middle level, we observe a separation between the models trained with reject bucket loss and HOD loss. This indicates that the difference in objective function provides the second most diversity.
At the lowest level, we observe all the $5$ vanilla ensemble models with same representation learning and loss function are clustered together. 
It is visible as 8 block matrices of size $5\times5$ along the diagonal of the heat matrix with lighter shades. This indicates that the diversity is minimum among them.

As detailed in Sec.~\ref{sec:ensemble}, we employ a greedy search algorithm over the pool of $40$ models to select our diverse ensemble members. 
The greedy search algorithm aims to select an ensemble member which provides the highest performance boost at every step.
This boost can be achieved by leveraging the diversity among the models using their complementary performance.
We believe that the selected diverse ensemble depends on the diversity of the pool of models.
To investigate this we employ the greedy search algorithm on different pools of models and plot the OOD AUROC for them at varying ensemble sizes in Fig.~\ref{fig:auc_vs_diveristy}.
The sub-pools were generated by the main diversity factors \ie{} representation learning and objective function as observed in Fig.~\ref{fig:heatmap_div}.
As a baseline we present the performance of the ImageNet pre-trained models with reject bucket (blue, pool size: 5).
We show the performance over pool of all models trained with reject bucket loss (purple, pool size: 20), pool of all models trained with HOD loss (red, pool size: 20), pool of all models pre-trained with contrastive learning (green, pool size: 20), pool of all models pre-trained with natural images (orange, pool size: 20) and overall pool (red, pool size: 40).
Identical models were selected over all the models and over all the models with HOD loss, and we observe this set of selected model to perform the best.
The selected diverse ensemble over pool of models pre-trained with natural images are better than the ones selected over the pool of contrastive pre-trained models at every ensemble size.
The diverse ensemble selected over models trained with reject bucket under-performs compared to the others.

Along with the selected ensemble members, it is also interesting to observe the order in which the greedy search algorithm selects them at every step. For example, the final diverse ensemble over all models has 3 BiT-L + HOD models and 2 MICLe + HOD models.
The sequence in which they were added to diverse ensemble is: BiT-L-HOD $\rightarrow$ MiCLe-HOD $\rightarrow$ BiT-L-HOD $\rightarrow$ MiCLe-HOD $\rightarrow$ BiT-L-HOD.
Note the alternate addition of BiT-L and MICLe models. This indicates that the greedy search algorithm attains the maximum boost in performance at every step by balancing the representational diversity of the models. 
The above results also suggest that introducing additional representation learning strategies and loss functions can further enrich the model pool diversity, and may lead to better OOD performance.

\subsection{Subgroup Analysis of OOD Detection Performance}
\label{sec:sub_group}
In this section, we investigate how our final diverse ensemble model performs across different skin condition risk categories and skin type subgroups. As a baseline model, we compare against a vanilla ensemble of $5$ ImageNet pre-trained models, trained with a reject bucket similar to~\cite{liu2020deep}. 

\begin{table*}[ht]
\centering
\caption{Subgroup analysis of OOD detection performance across different risk categories and skin type in the test set. For each of the subgroups we also indicate the inlier and outlier sample size used for computing AUROC. We compare our diverse ensemble model (3 BiT-L + HOD, 2 MICLe + HOD models) with baseline ensemble (5 ImageNet + reject bucket model). The best result score is indicated in \textbf{bold}.}
\vspace{0.5em}
\begin{tabular}{lcccc}
 \toprule
  Subgroups & Num. Inlier & Num. Outlier & Baseline ensemble &  Diverse ensemble \\
  & samples & samples & AUROC ($\uparrow$) & AUROC ($\uparrow$) \\
 \otoprule
 High Risk & $1192$ & $64$ & $81.6$ & $\mathbf{89.2}$ \\
 Medium Risk & $1192$ & $296$ & $77.9$ & $\mathbf{81.4}$ \\
 Low Risk & $1192$ & $577$ & $75.1$ & $\mathbf{83.3}$ \\
 \midrule
 Skin-types-1\&2 & $209$ & $127$  & $75.1$ & $\mathbf{82.9}$ \\
 Skin-types-3 & $364$ & $249$  & $78.2$ & $\mathbf{84.3}$ \\
 Skin-types-4 & $333$ & $215$  & $77.1$ & $\mathbf{84.8}$ \\
 Skin-types-5\&6 & $28$ & $18$ & $77.3$ & $\mathbf{94.3}$ \\
 \bottomrule
\end{tabular}
\label{tab:subgroup_analysis}
\end{table*}

\subsubsection{Analysis Over Risk Subgroups}
For this purpose, dermatologists labeled each of the $65$ outlier classes in the test set as one of the three risk categories. Low risk indicates conditions that may cause either no injury or only temporary discomfort; medium risk indicates conditions that can result in injury or impairment requiring medical intervention to prevent permanent impairment; high risk indicates conditions that can result in permanent impairment or life threatening injury or death. All risk category assignments are based on the worse case scenario if left untreated; for example, though conditions like `Urticaria' have a small chance to cause death, they are assigned to the high risk category.  
We present the results of this subgroup analysis on the test set in Table~\ref{tab:subgroup_analysis}. Each AUROC value compares the inliers \vs{} outliers of subgroup X. We observe that our diverse ensemble has a higher AUROC than the baseline ensemble by 7.6, 2.5 and 8.2 points for high-risk, medium-risk and low-risk subgroups, respectively. The most substantial gains are for the high-risk and low-risk subgroups.

\subsubsection{Analysis Over Skin Type Subgroups}
For this analysis we divide the test set samples based on Fitzpatrick skin types. We analyzed 4 subgroups: (i) Types-1\&2 (Pale-white and white skin), (ii) Type-3 (Beige skin), (iii) Type-4 (Brown skin) and (iv) Types-5\&6 (Dark brown and black skin). 
The objective of this analysis differs slightly from the risk subgroups. Instead of stratifying based on the category of skin condition (which a user may not know), here we analyze subgroups based on information that a user may know: their skin type.
For this purpose we computed our AUROC metric comparing inliers of subgroup X \vs{} outliers of subgroup X. We present the results on the test set in Table~\ref{tab:subgroup_analysis}.
Note that out of the $2129$ test samples, we had access to ground-truth skin types for $1543$ samples.
In general we observe that our diverse ensemble had a higher AUROC for all 4 subgroups. 
The margin of 17 points AUROC for Types-5\&6 is particularly striking as these skin types were rare in our dataset, though the smaller sample size render making confident conclusions difficult.

\begin{figure*}[t]
\centering
\includegraphics[width=.9\textwidth]{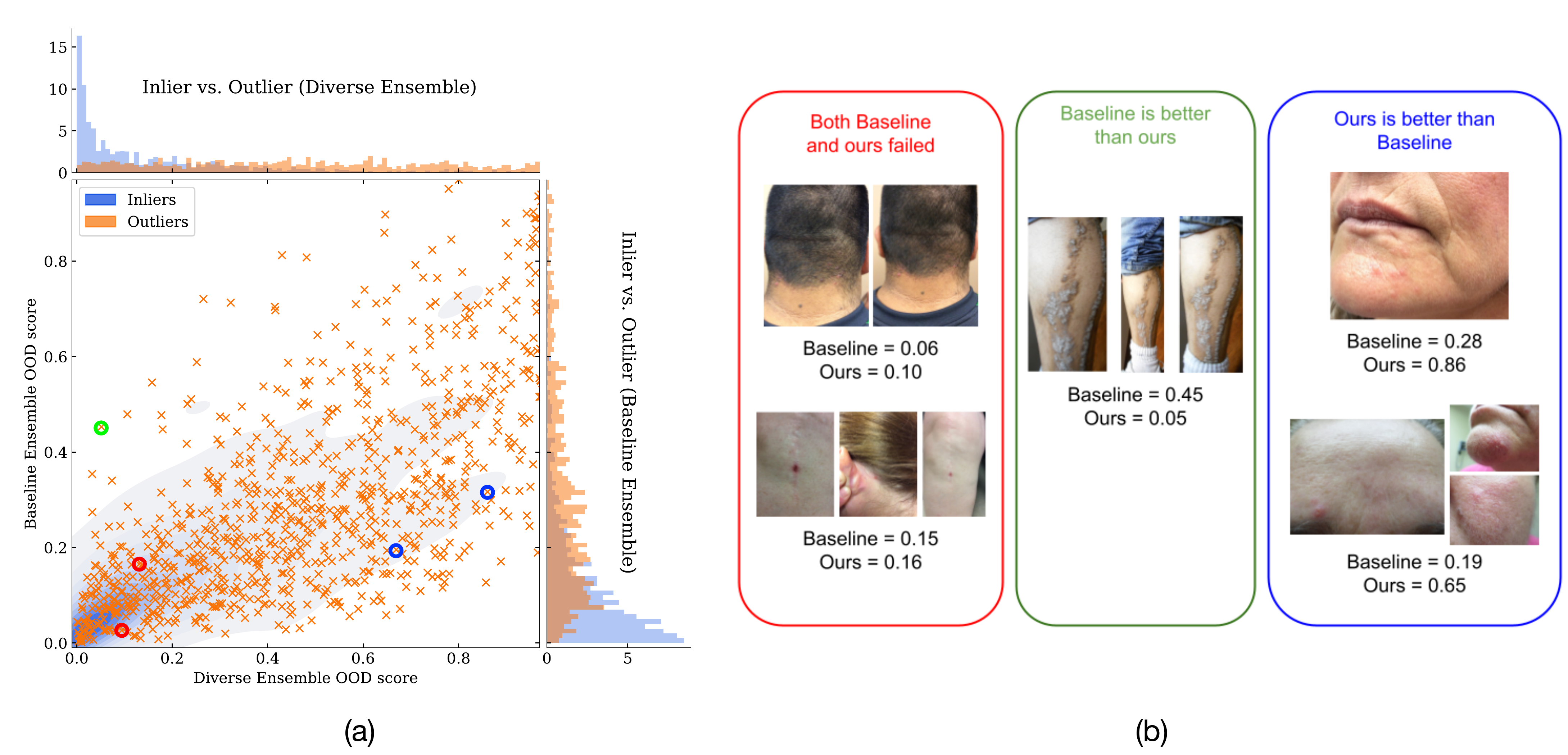}
\caption{Qualitative analysis of the results and failure cases for both our model and the baseline model. (a) This figure shows a scatter plot on the test data samples, with the X-axis being the OOD score $\mathcal{U}(\obs)$ assigned to the samples by our diverse ensemble and the Y-axis being the OOD score $\mathcal{U}(\obs)$ assigned to the samples by baseline ensemble. The inlier samples are shown in blue as a density plot. The outlier samples are shown in orange $\times$. The marginal distributions of inlier and outlier is also shown for baseline ensemble (right) and our diverse ensemble (top). (b) This figure shows some selected outlier samples from the test data for a qualitative analysis of the result. (Left) the red box shows two cases where both the baseline and our model failed to detect the outlier sample. The corresponding samples are shown in red circle in the scatter plot. (Middle) the green box shows a sample where baseline did better than our model. The corresponding sample are shown in green circle in the scatter plot. (Right) the blue box shows cases where our model did better than baseline. The corresponding samples are shown in blue circle in the scatter plot. The number of image instances differs across cases, and for each case we also present the OOD score $\mathcal{U}(\obs)$ assigned by both the baseline model and our model.}
\label{fig:qualitative}
\end{figure*}

\subsection{Qualitative Analysis and Failure Cases}
\label{sec:qualitative}

In this section, we perform a qualitative analysis of our results and show the failure cases. For this purpose, we compare our diverse ensemble with the baseline ensemble similar to Sec.~\ref{sec:sub_group}.
First we show a scatter plot (Fig.~\ref{fig:qualitative}a) of all the test outlier samples (in orange) and the inliner samples (in blue as a density plot in Fig.~\ref{fig:qualitative}a) comparing the OOD scores assigned by the baseline ensemble versus the diverse ensemble.
The outlier samples in the bottom-left region where inlier density is high corresponds to the difficult cases where both our model and baseline model fails. We select two cases from this category (indicated by red circle) and show them in Fig.~\ref{fig:qualitative}b (outlined in red box). 
The top-left region in Fig.~\ref{fig:qualitative}a represents the cases where baseline is better than our model. 
We select one such case (indicated by green circle) and show that in Fig.~\ref{fig:qualitative}b (outlined by green box). 
The bottom-right region in Fig.~\ref{fig:qualitative} represents the cases where our model outperformed the baseline model. 
We randomly select two cases from this region (indicated by blue circle in Fig.~\ref{fig:qualitative}a) and show them in Fig.~\ref{fig:qualitative}b (outlined by blue box). 
The top-right region in Fig.~\ref{fig:qualitative}a represents the cases where both the baseline and our models performed well. These mostly corresponds to relatively easy cases.

In Fig.~\ref{fig:qualitative}a, we also show the marginal distributions of inliers (indicated in blue) and outliers (indicated in orange) modelled by baseline ensemble (right) and by our diverse ensemble (top).
We can observe that the inlier distribution modelled by our diverse ensemble is more compact and narrower in comparison to the baseline ensemble. Also, we observe substantially more overlap between the inlier and outlier distributions for the baseline in comparison to our model. 
This provides qualitative evidence of our model being better at detecting outliers in comparison to the baseline.

\begin{figure}[h]
     \centering
     \begin{subfigure}[b]{0.235\textwidth}
         \centering
         \includegraphics[width=\textwidth]{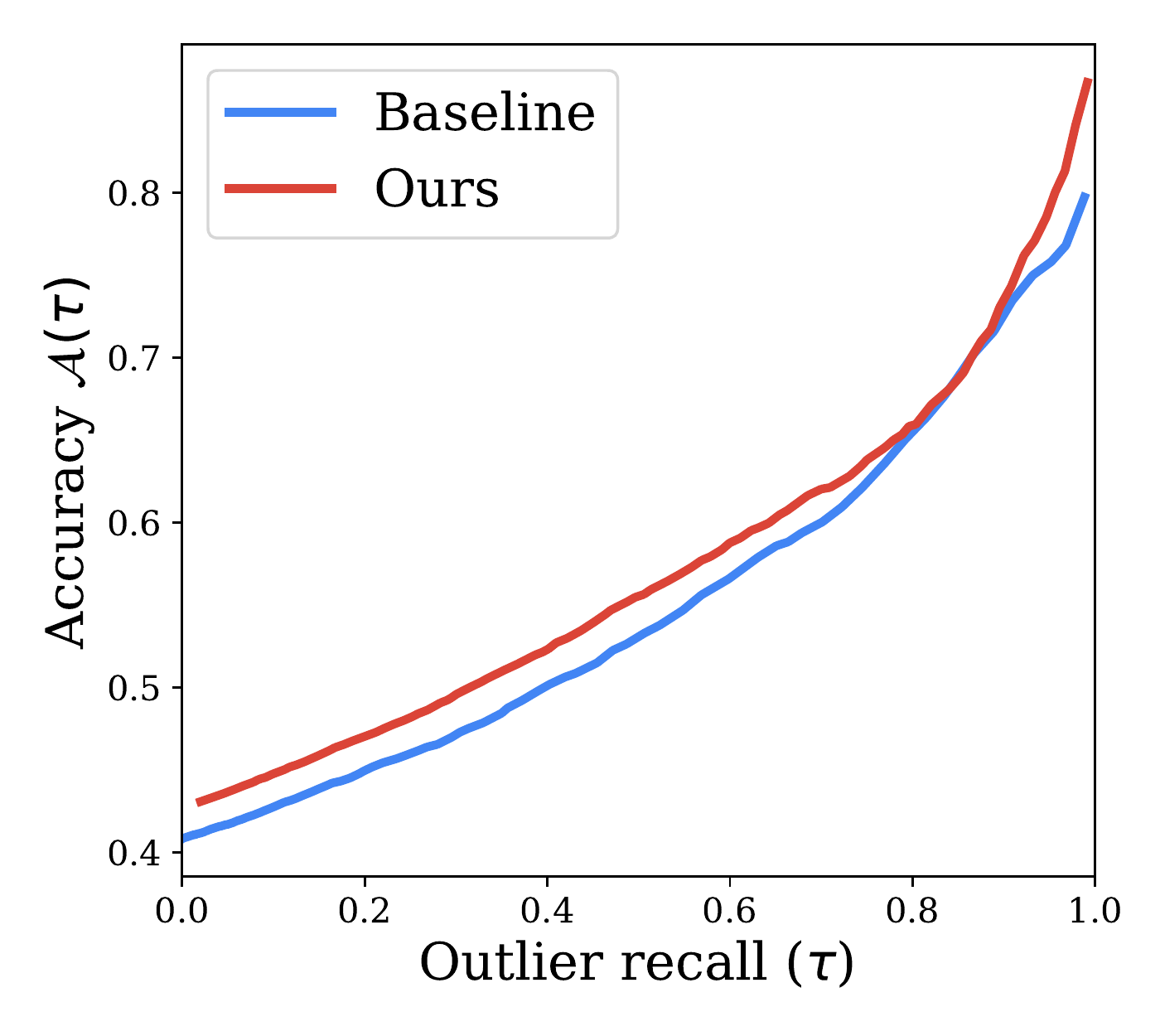}
         \caption{}
         \label{fig:conf_vs_acc}
     \end{subfigure}
     \hfill
     \begin{subfigure}[b]{0.235\textwidth}
         \centering
         \includegraphics[width=1.01\textwidth]{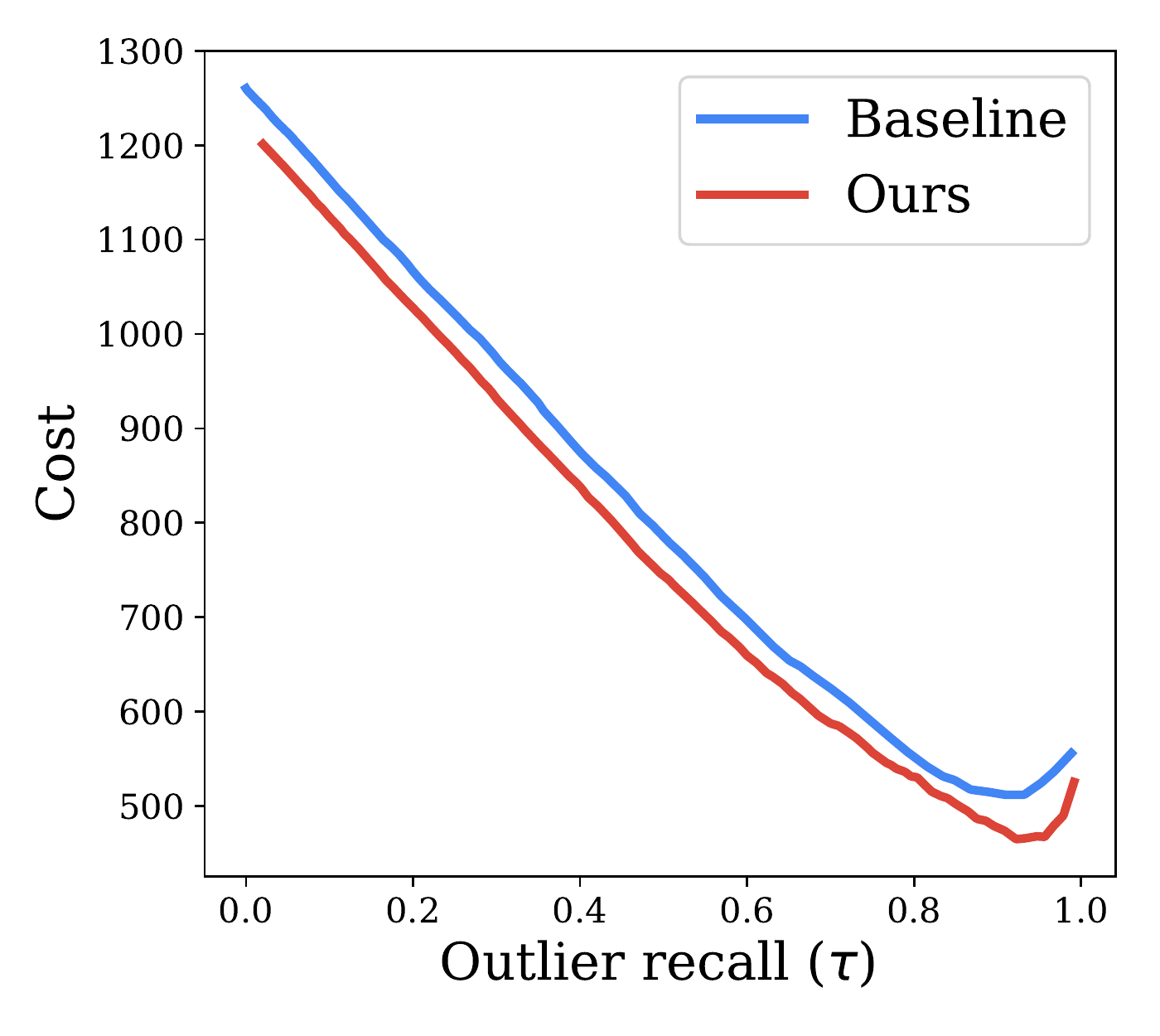}
         \caption{}
         \label{fig:conf_vs_cost}
     \end{subfigure}

        \caption{This figure shows the results of joint evaluation for inlier accuracy and outliers detection and model trust analysis. Our diverse ensemble model is indicated in red and the baseline ensemble model is indicated in blue. (a) Accuracy $\mathcal{A}(\tau)$ \vs{} outlier recall ($\tau$) is shown here at different confidence threshold $\tau$. $\mathcal{A}(\tau)$ is computed on test set as defined in Eqn.~\ref{eq:conf_vs_acc}.  
        (b) Cost ($\tau$) \vs{} outlier recall ($\tau$) curves for our diverse ensemble model and the baseline ensemble model. For different values of $\tau$, the cost is calculated based on the cost matrix indicated in Fig.~\ref{fig:cost} on the test set.}
        \label{fig:eval_ind_ood}
\end{figure}

\subsection{Joint Evaluation for Inliers and Outliers Predictions}

Although in this article we mainly focus on improving the performance for OOD detection, the ultimate goal for our dermatology model is to have a high detection accuracy for both inliers and outliers. 
Therefore, in this section we investigate the model's joint prediction accuracy for inliers and outliers. 

Let us consider a confidence score threshold $\tau$ above which we allow our model to predict, \ie{} if the test samples $\obs$ has a confidence scores $\mathcal{C(\obs)} > \tau$ its associated output will be predicted and if $\mathcal{C(\obs)} < \tau$ we abstain from prediction.
For a fixed $\tau$, corresponding model accuracy $\mathcal{A}(\tau$) for the predicted samples can be computed as
\begin{equation}
\label{eq:conf_vs_acc}
    \mathcal{A}(\tau) = \frac{\sum_{i} \mathbf{1}(\mathcal{C}(\obs_i)>\tau, \argmax_{c \in \SetOfInlierClasses} p(c|\obs_i)=\cls_i)}{\sum_{i} \mathbf{1}(\mathcal{C}(\obs_i)>\tau)},
\end{equation}
\noindent
which evaluates the inlier accuracy for the predicted samples, and considers all predictions for outliers incorrect.

Note that there is a calibration difference between our diverse ensemble model and the baseline ensemble model. Thus the $\mathcal{A}(\tau)$ of diverse ensemble and baseline ensemble for a fixed $\tau$ is not directly comparable.
Thus to normalize the confidence threshold $\tau$, we compute the outlier recall for the value of $\tau$, which adjusts for the differences in calibration of the models.

We evaluate the $\mathcal{A}(\tau)$ and outlier recall at different thresholds $\tau$ and plot them in Fig.~\ref{fig:conf_vs_acc} similar to~\citet{lakshminarayanan2016simple, van2020uncertainty}.
We observe that our diverse ensemble model has higher accuracy than the baseline ensemble model for all outlier recall rates. 
This indicates that our model is better at jointly identifying outliers and correct inliers in contrast to baseline for different choices of operating points.

\subsection{Model Trust Analysis}
\label{subsec:cost_analysis}
\begin{figure}[h]
\center{\includegraphics[width=.45\textwidth]{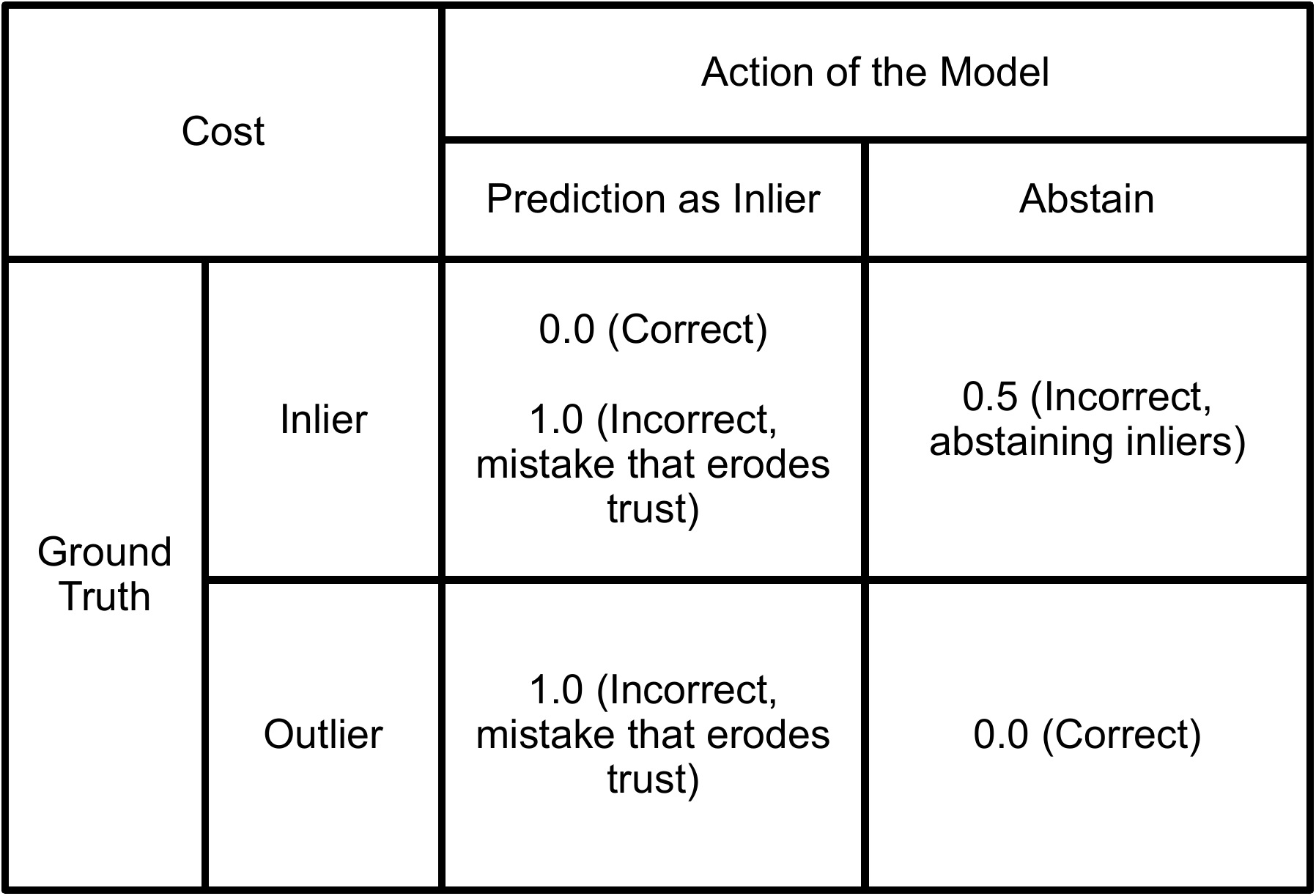}}
\caption{Cost matrix for all possible predictions for inliers and outliers used for model trust analysis.}
\label{fig:cost}
\end{figure}

In this section, we perform a model trust analysis to better understand the total downstream clinical implications of the model for misclassifying inliers and outliers. 
For a fixed confidence threshold $\tau$, we count the following types of mistakes: (i) incorrect prediction for inliers (\ie{} mistaking inlier condition A as inlier condition B), (ii) incorrect abstention of inliers (\ie{} abstaining from making a prediction for a inlier), (iii) incorrect prediction for outliers as one of the inlier classes. 
In order to account for the asymmetric clinical consequences of these different types of mistakes, we present a cost matrix assigning different costs for the different mistakes in Fig.~\ref{fig:cost}.
Both within-inlier incorrect predictions and outlier-as-inlier were penalized with a score of $1$. Such mistakes can potentially erode trust of the user in the model. 

Incorrect abstention of inliers as an outlier was penalized with a score of $0.5$ to reflect the fact that potential users of the model would be able to seek additional guidance given the model-expressed uncertainty or abstention.
Note that these numbers are qualitative approximations for modeling the downstream impact. Real-world scenarios are more complex and contain a variety of unknown variables; this has been simplified to better understand the main focus of this work: outlier detection and inlier accuracy. These choices were verified by a dermatologist.

We estimate the cost for both our diverse ensemble model and baseline ensemble model at different values of confidence thresholds $\tau$ and show the plot in Fig.~\ref{fig:conf_vs_cost}. 
Similar to the accuracy, we use outlier recall at different $\tau$ to adjust for the different calibrations of the $\tau$ in the two models.
We observe that our diverse ensemble model has lower overall cost than the baseline ensemble model across all outlier recall rates, and the lowest cost is achieved at the outlier recall rate $=0.92$. The consistently lower cost indicates our model performs better after accounting for different downstream clinical implications.

\section{Conclusion}
\label{sec:conc}

In real-world deployment of medical machine learning models, test inputs with previously unseen conditions are often encountered.  
For safety it may be important to identify such inputs and abstain from classification, in order to guard against inappropriate over-reliance on model outputs and empower model users to pursue safe next steps such as consulting a clinician.

In this article, we tackle this challenging task of detecting the previously unseen long-tail of rare conditions for a dermatology classification model. We frame this task as an out-of-distribution (OOD) detection problem.
Leveraging our labeled dataset with a long-tail of conditions, we first construct a benchmark for reliably evaluating OOD methods. 

We further propose a novel hierarchical outlier detection (HOD) loss function for OOD detection. In contrast to existing approaches of assigning a single abstention class for OOD, we assign multiple abstention classes corresponding to the number of OOD classes available in the train set. Also, in addition to the fine-grained classification loss, we include a coarse loss to aid high level clustering of inliers and outliers. We demonstrate the additional performance gains from using HOD loss compared to using baselines.

We additionally explore the utility of the HOD loss in the context of multiple different state-of-the-art representation learning methods. Beyond the commonly used ImageNet pre-trained model, we investigate the BiT model pre-trained on large-scale JFT dataset, and contrastive pre-training based SimCLR and MICLe approaches. We demonstrated that better representation learning can improve OOD detection performance, and that these can also be improved via the HOD loss.

We also explored different ensembling strategies.
A vanilla ensemble improved both the OOD performance and inlier accuracy for all the models trained with and without HOD loss and different representation learning techniques.
The diverse ensemble selection approach using a greedy search algorithm on a pool of models with different representation learning and loss functions further outperformed the vanilla ensemble models both for OOD performance and inlier accuracy.

We also investigated the OOD performance for different subgroups of risk levels and skin types. We show that our proposed method demonstrated superior performance in comparison to the baseline for all the subgroups.

To quantify potential downstream clinical implications, we go beyond the traditional performance metrics and construct a cost matrix for model trust analysis.
We demonstrate the superiority of our method over baseline in this metric, indicating the effectiveness of our model for real-world deployment. 
Ideally we may directly optimize for this cost matrix as an objective function during training. We leave this as a possible future work.


All in all, we believe that our proposed approach can aid successful translation of AI algorithms into real-world scenarios. Although we have primarily focused on OOD detection for dermatology, most of our contributions are fairly generic and can be easily to generalized to OOD detection in other applications.

\section*{Acknowledgements}
We would like to thank Olaf Ronneberger, Ali Eslami, Rudy Bunel, Simon Kohl, Krishnamurthy Dvijotham from DeepMind, Jonathan Deaton from Google Health and Neil Houlsby from Google Research for helpful discussions towards initial ideation. We would also like to express our appreciation towards Joshua Dillon, Jasper Snoek, Jeremiah Liu from Google Research, and Atilla Kiraly, Terry Spitz, and Dale Webster from Google Health, and Kimberly Kanada for insightful discussion and providing valuable feedback for this work, and Jay Hartford from Google Health for assistance in data-related logistics.

\bibliographystyle{model2-names.bst}\biboptions{authoryear}
\bibliography{main.bbl}


\begin{table*}[t]
\centering
\scriptsize
\caption{List of dermatological conditions in our dataset.}
\begin{tabular}{p{1in}p{5in}}
 \toprule
 Inlier conditions & Acne, Tinea, Tinea Versicolor, Actinic Keratosis, Folliculitis, Allergic Contact Dermatitis, Alopecia Areata, Urticaria, Androgenetic Alopecia, Verruca vulgaris, Vitiligo, Basal Cell Carcinoma, Hidradenitis, Post-Inflammatory hyperpigmentation, Lentigo, Psoriasis, Cyst, SCC/SCCIS, SK/ISK, Scar Condition, Seborrheic Dermatitis, Melanocytic Nevus, Melanoma, Skin Tag, Eczema, Stasis Dermatitis \\
 \midrule
 Outlier conditions in Train set & Abscess, Erythema annulare centrifugum, Acanthosis nigricans, Erythema dyschromicum perstans, Erythema gyratum repens, Tattoo, Traction alopecia, Traumatic ulcer, Acute generalised exanthematous pustulosis, Nevus sebaceous, Nevus spilus, Fordyce spots, Ochronosis, Foreign body reaction of the skin, Amyloidosis of skin, Varicose veins of lower extremity, Ganglion cyst, Viral Exanthem, Xanthoma, Angiosarcoma of skin, Oral fibroma, Yellow nail syndrome, Hairy tongue, Atypical Nevus, Pearly penile papules, Pemphigus foliaceus, Pemphigus vulgaris, Blue sacral spot, Hordeolum internum, Photodermatitis, Phrynoderma, Burn of skin, Calcinosis cutis, Infected eczema, Pitted keratolysis, Insect Bite, Chilblain, Poikiloderma, Irritant Contact Dermatitis, Porphyria cutanea tarda, Keratolysis exfoliativa, Keratosis pilaris, Pretibial myxedema, Prurigo nodularis, Pruritic urticarial papules and plaques of pregnancy, Leukonychia, Cutaneous capillary malformation, Lichen Simplex Chronicus, Lichen nitidus, Lichen planopilaris, Cutaneous lupus, Lichen planus/lichenoid eruption, Cutaneous metastasis, Puncture wound - injury, Lichenoid myxedema, Purpura, Cutaneous sarcoidosis, Lipodermatosclerosis, Pyoderma Gangrenosum, Livedo reticularis, Livedoid vasculopathy, Dental fistula, Diabetic dermopathy, Diabetic ulcer, Dissecting cellulitis of scalp, Drug Rash, Skin striae \\
 \midrule
 Outlier conditions in Validation set & Necrobiosis lipoidica, Telangiectasia disorder, Telogen effluvium, Erythema nodosum, Erythrasma, Symmetrical dyschromatosis of extremities, Nevus lipomatosus cutaneous superficialis, Traumatic bulla, Nevus of Ota, Trichotillomania, Onycholysis, Onychomadesis, Venous Stasis Ulcer, Onychomycosis, Onychoschizia, Granuloma annulare, Granulomatous cheilitis, Hailey Hailey disease, Paronychia, Hematoma of skin, Pemphigoid gestationis, Becker’s nevus, Benign neoplasm of nail apparatus, Hemosiderin pigmentation of skin, Hirsutism, Periungual fibroma, Perleche, Hypersensitivity, Bullosis diabeticorum, Idiopathic guttate hypomelanosis, Pigmented purpuric eruption, Impetigo, Pincer nail deformity, Candida, Canker sore, Cellulitis, Central centrifugal cicatricial alopecia, Pityriasis lichenoides, Pityriasis rosea, Chondrodermatitis nodularis, Porokeratosis, Clubbing of fingers, Post-Inflammatory hypopigmentation, Knuckle pads, Condyloma acuminatum, Confluent and reticulate papillomatosis, Leprosy, Leukocytoclastic Vasculitis, Cutaneous T Cell Lymphoma, Pseudopelade, Pterygium of nail, Lichen sclerosus, Lichen striatus, Cutaneous neurofibroma, Cylindroma of skin, Deep fungal infection, Dermatitis herpetiformis, Dermatofibroma, SJS/TEN, Dermatomyositis, Scabies, Morphea/Scleroderma, Melasma, Molluscum Contagiosum, Erosive pustular dermatosis, Erythema ab igne \\
 \midrule
 Outlier conditions in Test set & Nail dystrophy due to trauma, Syphilis, Accessory nipple, Acne keloidalis, Acquired digital fibrokeratoma, Erythema migrans, Erythema multiforme, Erythromelalgia, Nevus comedonicus, Trachyonychia, Triangular alopecia, Adnexal neoplasm, Folliculitis decalvans, Notalgia paresthetica, O/E - ecchymoses present, Fox-Fordyce disease, Angiofibroma, Angiokeratoma of skin, Onychorrhexis, Xerosis, Osteoarthritis, Zoon’s balanitis, Arsenical keratosis, Grover’s disease, Hand foot and mouth disease, Hemangioma, Beau's lines, Herpes Simplex, Herpes Zoster, Perforating dermatosis, Perioral Dermatitis, Hyperhidrosis, Brachioradial pruritus, Ichthyosis, Bullous Pemphigoid, Cafe au lait macule, Pilonidal cyst, Inflammatory linear verrucous epidermal nevus, Inflicted skin lesions, Ingrown hair, Pityriasis alba, Pityriasis amiantacea, Chicken pox exanthem, Intertrigo, Pityriasis rubra pilaris, Clavus, Kaposi's sarcoma of skin, Keratoderma, Pressure ulcer, Comedone, Leukoplakia of skin, Pseudolymphoma, Pyogenic granuloma, Lipoma, Cutis verticis gyrata, Retention hyperkeratosis, Rheumatoid nodule, Longitudinal melanonychia, Rosacea, Dermatofibrosarcoma protuberans, Sebaceous hyperplasia, Skin and soft tissue atypical mycobacterial infection, Milia, Epidermal nevus, Mucocele \\
 \bottomrule
\end{tabular}
\label{tab:list_of_cond}
\end{table*}

\end{document}